\documentclass{article} % For LaTeX2e
\usepackage{iclr2027_conference,times}

\usepackage{amsmath,amsfonts,bm}

\def\eqref#1{equation~\ref{#1}}
\def\1{\bm{1}}

\DeclareMathAlphabet{\mathsfit}{\encodingdefault}{\sfdefault}{m}{sl}
\SetMathAlphabet{\mathsfit}{bold}{\encodingdefault}{\sfdefault}{bx}{n}

\usepackage{hyperref}
\usepackage{url}
\usepackage{graphicx}
\usepackage{booktabs}
\usepackage{placeins}
\usepackage{capt-of}
\usepackage{wrapfig}
\usepackage{needspace}
\usepackage[inline]{enumitem}
\usepackage{amsthm}
\newtheorem{proposition}{Proposition}
\usepackage{tcolorbox}
\tcbuselibrary{breakable,skins}
\definecolor{propositionred}{RGB}{210,15,22}
\tcolorboxenvironment{proposition}{enhanced jigsaw,breakable,
  colback=propositionred!4!white,
  colframe=propositionred,boxrule=0.6pt,arc=1.5pt,
  left=5pt,right=5pt,top=3pt,bottom=3pt,
  before upper={\displaywidowpenalty=10000\brokenpenalty=10000},
  before skip=10pt,after skip=10pt}

\title{Reactive Real-Time Flow Policies via\\ Asynchronous Distribution Alignment}

\author{Moritz Zoellner\textsuperscript{1,2},\quad
Reece O'Mahoney\textsuperscript{2},\quad
Ioannis Havoutis\textsuperscript{2},\quad
Rohan Paleja\textsuperscript{1}\\[5pt]
\normalfont\fontencoding{T1}\fontsize{8}{10}\selectfont

\begin{tabular}{@{}l@{\quad}r@{}}
\textsuperscript{1}\ \texttt{Department of Computer Science, Purdue University} &
{\fontfamily{pcr}\selectfont\{zoellner, rpaleja\}@purdue.edu}\\[2pt]
\textsuperscript{2}\ \texttt{Oxford Robotics Institute, University of Oxford} &
{\fontfamily{pcr}\selectfont\{reeceo,ioannis\}@robots.ox.ac.uk}
\end{tabular}
}

\iclrfinalcopy % Show authors and disable review line numbers.
\begin{document}

\maketitle
\pagestyle{plain} % Neutral preprint: page numbers without the ICLR header.

\begin{abstract}
Generalist robot policies such as vision-language-action models (VLAs) have achieved remarkable generalization, but their inference delays can conflict with the demands of real-time control. Asynchronous execution avoids pauses between action chunks by predicting the next sequence of actions while the robot carries out the previous one. In this paper, we study whether asynchronous execution produces the same action distribution as the original VLA. We find that, for non-Markovian demonstrations, asynchronous execution can produce a fundamentally different action distribution, which can limit the policy’s reactivity. In our method, we seek to restore this reactivity by aligning the asynchronously produced action distribution with that of the original VLA through two complementary mechanisms. First, \emph{Recursive Flow-Field Distillation} trains the asynchronous policy using the VLA’s action-generation flow. We characterize the learned distribution theoretically and show experimentally that our asynchronous policy can generate nearly the full range of actions the original VLA would produce, while existing asynchronous methods recover only a fraction of that range. Second, \emph{Propose–Resolve} prepares multiple action sequences asynchronously and uses the latest observation to select among them based on a lightweight approximation of their likelihood under the VLA’s action distribution. Our resulting method matches the original VLA’s success on LIBERO and retains about 80\% of its success on RoboMimic, about 30 percentage points more than existing asynchronous methods.
\end{abstract}

\section{Introduction}
\label{sec:introduction}

Generalist policies, such as Vision-language-action (VLA) models, are becoming better at learning general-purpose behavior from large, diverse datasets~\citep{kim2024openvla,black2024pi0,shukor2025smolvla}. Yet the inference times of these large models limit real-time control: robots may need to dispatch actions at a much higher rate than the policy can produce them. \emph{Action chunking} partly mitigates this mismatch by predicting multiple future actions at once, at the cost of reactivity~\citep{zhao2023act,chi2023diffusionpolicy}. Under \emph{synchronous execution}, however, the robot must still wait between chunks while the next prediction is generated. While such pauses may be acceptable for some tasks, they limit throughput and become inherently detrimental in continuous or dynamic settings~\citep{black2025rtc}.

\emph{Asynchronous execution} addresses this limitation by starting computation of the next action chunk before execution of the current chunk has finished, such that inference overlaps with ongoing control~\citep{black2025rtc,black2025trainingtimertc,ho2026paint}. Real-Time Chunking (RTC) enables this by matching the overlapping portions of consecutive chunks~\citep{black2025rtc}. Its central insight is that asynchronous execution remains reliable only when the new prediction is compatible with the actions already committed by the preceding chunk.
This original approach to continuous execution, however, introduces a new mismatch: because the next chunk is generated before the current one has finished executing, it is conditioned on an earlier state. During inference, the robot continues moving, so that at \emph{handoff}, when execution passes to the newly generated chunk, the robot will already be in a different state from the one on which that chunk was conditioned. This can limit reactivity and reduce performance~\citep{black2025rtc,park2026pir2}.

\begin{figure}[t]
  \centering
  \includegraphics[width=\textwidth]{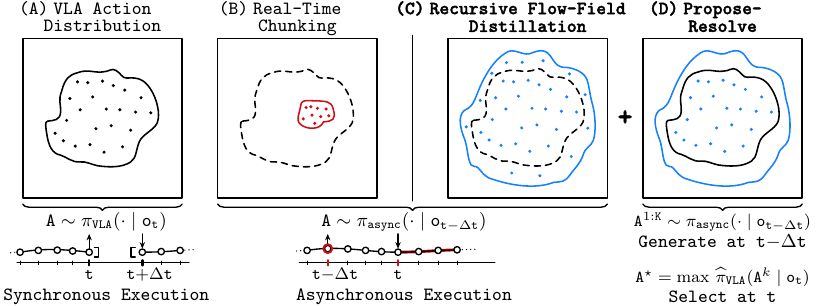}
  \caption{\textbf{Asynchronous execution and distribution alignment.}
  The point clouds schematically represent different action distributions at $t$, with each point corresponding to a possible action sample; inference of such a sample takes $\Delta t$.
  \emph{(A) Synchronous execution} queries the VLA at time $t$, so execution has to wait until the action is generated at $t+\Delta t$.
  \emph{(B--C) Asynchronous execution} instead starts generation at $t-\Delta t$, while the robot carries out its committed actions, so that the next chunk is prepared for execution at $t$. \emph{(B) Real-Time Chunking} produces a narrower subset of the actions compared to the VLA when the training data is non-Markovian.
  \emph{(C) Recursive Flow-Field Distillation} trains the asynchronous policy using the VLA's action-generation flow, recovering a broader proposal distribution.
  \emph{(D) Propose--Resolve} uses the latest observation to estimate each candidate's likelihood under the corresponding VLA distribution and selects the highest-scoring continuation.}
  \label{fig:intro-overview}
\end{figure}

To investigate this mismatch, we study the relationship between two possible action chunks at the same handoff state: one produced asynchronously by RTC, and the other by querying the VLA synchronously. We find that, even when both policies are learned from the exact same demonstrations, non-Markovian data can cause asynchronous execution methods to produce a fundamentally different action distribution from the VLA. Figure~\ref{fig:intro-overview}(A--B) exemplifies how RTC, while removing the inference delay, might produce only a narrow subset of the actions that the VLA would generate.

In our method, we seek to recover the original VLA’s action distribution asynchronously to restore the reactivity that current asynchronous execution methods lose. To avoid inheriting the history dependence of non-Markovian demonstrations, we first change what the asynchronous policy is trained to predict. Instead of supervising the asynchronous policy with the continuation recorded in the demonstration, we train it to reproduce the VLA’s behavior after the already-committed actions have been executed. We realize this objective through \emph{Recursive Flow-Field Distillation} (RFD), transferring the VLA’s action-generation flow into the asynchronous policy. When the future observation is uncertain, this learning target marginalizes over the VLA action distributions associated with the possible observations that could be reached. As illustrated in Figure~\ref{fig:intro-overview}(C), this can produce a broader proposal distribution than the VLA distribution associated with the observation ultimately reached.
The broader proposal distribution must therefore be aligned with the distribution the VLA would produce once the actual observation becomes available. To this end, \textit{Propose--Resolve} proposes multiple candidate actions asynchronously and resolves among them once the observation at handoff arrives. A lightweight resolver fits a diagonal-Gaussian approximation to the VLA distribution conditioned on the reached observation, uses it to estimate each candidate’s likelihood, and selects the highest-scoring continuation. This exploits a computational asymmetry: generating a new action sequence requires expensive iterative flow denoising, whereas evaluating a likelihood of an already-generated candidate can be a much cheaper operation. Figure~\ref{fig:intro-overview}(D) illustrates how this second step filters the broader proposal set toward the VLA distribution (A), aligning the asynchronously generated behavior with what the VLA would produce from the reached observation.

In summary, our method moves the expensive generative computation into the past, while still executing a reactive action that the VLA itself could produce from the observation available in the present. Our experiments show that the new learning objective recovers nearly the full range of actions the original VLA would produce, while existing asynchronous methods recover only a fraction of that range. Together with Propose--Resolve, we evaluate this approach in closed-loop execution: our complete method matches the original VLA’s success on LIBERO~\citep{liu2023libero} and retains about 80\% of its success on RoboMimic~\citep{mandlekar2022robomimic}, approximately 30 percentage points more than existing asynchronous methods. Interestingly, our ablations show that closer distributional alignment of RFD is substantially more important on RoboMimic than on LIBERO, suggesting that the value of reactivity versus continuation coherence is task dependent.

\FloatBarrier
\section{Preliminaries and Problem Setting}
\label{sec:handoff}

\noindent\textbf{Synchronous execution.}
We first consider the standard execution of an action-chunking VLA. Given an observation $o_t$, the policy generates an $H$-step action chunk $B_t \sim \pi_{\mathrm{VLA}}(\cdot \mid o_t)$. Only the first $s$ actions, $A_t := B_t[0:s]$, are executed before the policy is queried again. By slight abuse of notation, we also write $A_t \sim \pi_{\mathrm{VLA}}(\cdot\mid o_t)$ for the induced marginal over this executed segment. Under synchronous execution, the robot must wait at time $t$ for the VLA to finish inference of $A_t$.

\begin{figure}[!b]
\centering
\includegraphics{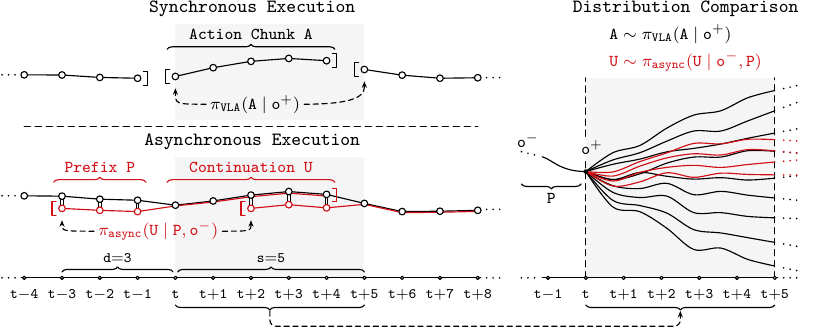}
\caption{\textbf{Execution timing and action distributions at handoff.}
Schematic with $d=3$ and $s=5$. \textbf{Left:} \emph{Synchronous execution} queries the VLA at time $t$ for the five actions executed over $[t,t+s]$. Asynchronous execution instead begins generation at $t-d$: the first $d$ actions form the committed prefix $P$, while the following $s$ actions form the continuation $U$ executed after handoff. Note that the schematic assumes prefix-compatible generation, so $U$ begins from the state reached after executing $P$; enforcing this compatibility is the original objective from RTC.
\textbf{Right:} The resulting distributions over the same $s$-step execution interval. Black curves show possible VLA action chunks conditioned on $o^+$, while red curves show asynchronous continuations generated from $(o^-,P)$.}
\label{fig:rtc-execution}
\label{fig:handoff}
\end{figure}
\noindent\textbf{Asynchronous execution.}
To make the next action chunk available without waiting at time $t$, asynchronous execution starts its inference $d$ control steps earlier, at $t-d$, while the robot continues executing the current plan. For example, $150$\,ms of inference at $20$\,Hz (i.e. actions required every 50ms) corresponds to $d=3$ control steps of advance preparation. The actions that execute during this inference window are already committed; we collect them in the prefix $P_t := (a_{t-d},\ldots,a_{t-1})$. Asynchronous methods use this known prefix to condition or constrain generation of a new $H$-step chunk. Let $\smash{\widetilde{B}_{t-d} \sim \pi_{\mathrm{async}}(\cdot\mid o_{t-d},P_t)}$ denote the chunk produced by this asynchronous generation procedure. The next $s$ actions used after handoff are $\smash{U_t := \widetilde{B}_{t-d}[d:d+s]}$, which we call the executed \emph{continuation}. The resulting execution plan can therefore be written as $[\,P_t;U_t;\ldots\,]$: $P_t$ is fixed by the preceding plan, while $U_t$ is newly generated for execution after handoff. Again, by slight abuse of notation, we write $U_t \sim \pi_{\mathrm{async}}(\cdot\mid o_{t-d},P_t)$ for the continuation distribution induced by the complete asynchronous generation procedure.

For ease of notation, we henceforth consider a single handoff at time $t$ and omit its subscript, writing $A:=A_t$, $U:=U_t$, and $P:=P_t$. We denote the observation available when generation begins by $o^-:=o_{t-d}$, and the observation available at handoff by $o^+:=o_t$. Figure~\ref{fig:rtc-execution} (left) summarizes the asynchronous generation timeline and the corresponding variables introduced above. We thus obtain two action distributions for the same $s$-step interval beginning at $o^+$:
\begin{equation}
A \sim \pi_{\mathrm{VLA}}(\cdot \mid o^+),
\qquad
U \sim \pi_{\mathrm{async}}(\cdot \mid o^-,P).
\label{eq:handoff-distributions}
\end{equation}
Figure~\ref{fig:rtc-execution} (right) illustrates these two distributions over the same execution interval: the VLA distribution conditioned on $o^+$, and the asynchronous distribution generated based on $(o^-,P)$. At time $t$, the asynchronous continuation is already available, whereas obtaining a new VLA sample would require another inference. This motivates the central question of our analysis: \emph{how does the action distribution produced by asynchronous execution compare with that of the original VLA?}

\section{Distributional Analysis}
\label{sec:theory}

Train-Time Real-Time Chunking (TT-RTC) explicitly learns the asynchronous continuation distribution from demonstrations~\citep{black2025trainingtimertc}, giving us a direct setting in which to compare it with the original VLA. Consider one demonstrated segment $(o^-,P,o^+,U_{\mathcal D}) \sim \mathcal D$, corresponding to a trajectory in which the committed prefix $P$ carries the system from $o^-$ to $o^+$, followed by continuation $U_{\mathcal D}$. Across the demonstration distribution $\mathcal D$, this defines two natural prediction problems and their population targets:
\begin{equation}
\begin{array}{rcl@{\qquad\qquad}rcl}
(o^-,P) & \longrightarrow & U_{\mathcal D}
&
\pi_{\mathrm{TT\text{-}RTC}}(U\mid o^-,P)
& \approx &
p_{\mathcal D}(U\mid o^-,P)
\\[3pt]
o^+ & \longrightarrow & U_{\mathcal D}
&
\pi_{\mathrm{VLA}}(A\mid o^+)
& \approx &
p_{\mathcal D}(U\mid o^+)
\end{array}
\label{eq:two-prediction-problems}
\end{equation}
Both policies are therefore trained from the same demonstrations to predict the same future action segment; only the point in the trajectory on which that prediction is conditioned differs. Assuming a fully observed system with deterministic dynamics, the handoff observation is itself determined by the earlier observation and committed prefix, $o^+=f(o^-,P)$. Since there is no uncertainty about the state reached at handoff, one might expect predicting the continuation from $(o^-,P)$ to recover the same distribution as querying the policy at $o^+$. This formally captures our central analysis question at the population level: $p_{\mathcal D}(U\mid o^-,P)\mathrel{\smash{\stackrel{?}{=}}}p_{\mathcal D}(U\mid o^+)$. Proposition~\ref{prop:continuation-gap} characterizes its answer.

\begin{proposition}[Continuation mismatch]
\label{prop:continuation-gap}
Let $O^-$ and $O^+$ denote the random observations at inference time and handoff under $\mathcal D$; $P$ and $U$ denote the committed prefix and subsequent continuation. Assume fully observed deterministic dynamics, such that $O^+ = f(O^-,P)$. Then
\begin{equation}
\mathbb E_{O^-,P}\!\left[
D_{\mathrm{KL}}\!\left(
p_{\mathcal D}(U\mid O^-,P)\,\Vert\,
p_{\mathcal D}(U\mid O^+)
\right)
\right]
= I_{\mathcal D}(U;O^-,P\mid O^+).
\label{eq:continuation-gap}
\end{equation}
Consequently, the two population continuation targets coincide almost everywhere if and only if the demonstrated continuation satisfies the Markov property with respect to the handoff observation:
\begin{equation}
p_{\mathcal D}(U\mid O^-,P)
= p_{\mathcal D}(U\mid O^+)
\quad\text{a.e.}
\qquad\Longleftrightarrow\qquad
U \perp (O^-,P)\mid O^+.
\label{eq:continuation-equivalence}
\end{equation}
That is, once $O^+$ is known, the earlier observation $O^-$ and committed prefix $P$ provide no additional information about the continuation $U$.
\end{proposition}

Surprisingly, deterministic dynamics alone are not sufficient for the two continuation distributions to coincide. Recent work studying robotic imitation learning finds that this property does not generally hold in human demonstrations~\citep{lazzati2026chunking,zeng2026openloop}. For such non-Markovian data, even a perfectly learned $\pi_{\mathrm{TT\text{-}RTC}}$ will generally produce a different action distribution from the VLA. In our method, we therefore study how $\pi_{\mathrm{async}}$ can be aligned directly with $\pi_{\mathrm{VLA}}$ to preserve the VLA’s range of possible responses. Appendix~\ref{app:target-gap} provides the proof of Proposition~\ref{prop:continuation-gap} and discusses how the argument extends beyond deterministic dynamics and full observability.

\section{Method}
\label{sec:method}

Starting from a base policy $\pi_{\mathrm{VLA}}$ trained on demonstrations $\mathcal D$, we use its learned action distribution as the reference for constructing $\pi_{\mathrm{async}}$. We exploit this reference at two stages of asynchronous execution. First, for the continuation generated at $o^-$, we change what the asynchronous policy is trained to predict by aligning its output with the VLA distribution that will be relevant at handoff. Second, once $o^+$ becomes available, we use the VLA distribution conditioned on that observation to choose among continuations generated earlier. We realize these two interventions through \textbf{Recursive Flow-Field Distillation} and \textbf{Propose--Resolve}, respectively.

\subsection{Recursive Flow-Field Distillation}
\label{sec:flow-field-distillation}

Because the training objectives below operate on full $H$-step action chunks, we write each demonstrated transition as $(o^-,P,o^+,B_{\mathcal D})$, where $U_{\mathcal D}=B_{\mathcal D}[d:d+s]$ is the executed continuation analyzed in Section~\ref{sec:theory}. We retain the conditioning $(o^-,P)$, but replace the demonstrated future actions with supervision from the frozen VLA conditioned on $o^+$. A direct implementation, which we call \textit{teacher-sample supervision}, samples a full teacher chunk $B_V\sim\pi_{\mathrm{VLA}}(\cdot\mid o^+)$ and uses $(o^-,P,B_{V,0:H-d})$ as an ordinary continuation-training example. However, this exposes the student to the teacher's generative distribution only through sampled action chunks. Because both teacher and student are flow policies, we can instead transfer the teacher's vector field directly.

Given a teacher chunk $B_V\sim\pi_{\mathrm{VLA}}(\cdot\mid o^+)$, we sample noise $\epsilon$ and a flow time $\tau$ and construct the intermediate flow state $z_\tau=\tau\epsilon+(1-\tau)B_V$. For the asynchronous policy, we shift the first $H-d$ positions of this intermediate state behind the committed prefix, yielding $\bar z_\tau=[P;z_{\tau,0:H-d}]$. We denote the resulting asynchronous policy by $\pi_{\mathrm{RFD}}$ and its velocity field by $v_{\mathrm{RFD}}$. We then match its continuation velocities at positions $d{:}H$ to the VLA's velocities at positions $0{:}H-d$:
\begin{equation}
\mathcal L_{\mathrm{RFD}}
=
\mathbb E_{B_V,\epsilon,\tau}
\left[
\operatorname{MSE}\!\left(
v_{\mathrm{RFD}}(\bar z_\tau,\tau\mid o^-)_{d:H},
v_{\mathrm{VLA}}(z_\tau,\tau\mid o^+)_{0:H-d}
\right)
\right].
\end{equation}
We call this procedure \textit{recursive} because the VLA learned from the demonstrations is itself fed back into the learning process: its learned flow field provides the supervision used to train $\pi_{\mathrm{RFD}}$. Proposition~\ref{prop:rfd-population-target} characterizes the population continuation distribution induced by $\mathcal L_{\mathrm{RFD}}$. Appendix~\ref{app:teacher-supervision} provides the proof of Proposition~\ref{prop:rfd-population-target} and compares RFD with teacher-sample supervision.

\begin{proposition}[Target of $\pi_{\mathrm{RFD}}$]
\label{prop:rfd-population-target}
Let $\mathcal D$ induce any joint distribution over $(O^-,P,O^+)$. Assume that the frozen VLA provides the exact flow-matching field for the interpolation used in training, that $\pi_{\mathrm{RFD}}$ reaches the population optimum of the RFD objective, and that its flow is integrated exactly. Then the induced distribution over the executed continuation is
\begin{equation}
\pi_{\mathrm{RFD}}^\star(\cdot\mid o^-,P)
=
\mathbb E_{O^+\sim p_{\mathcal D}(\cdot\mid o^-,P)}
\left[
\pi_{\mathrm{VLA}}(\cdot\mid O^+)
\right].
\label{eq:rfd-population-target}
\end{equation}
\end{proposition}
Under the assumptions of Proposition~\ref{prop:continuation-gap}---full observability and deterministic dynamics, such that $(o^-,P)$ uniquely determines $o^+$---Equation~(\ref{eq:rfd-population-target}) reduces to $\pi_{\mathrm{RFD}}^\star(\cdot\mid o^-,P)=\pi_{\mathrm{VLA}}(\cdot\mid o^+)$. Thus, RFD removes the continuation mismatch identified in Section~\ref{sec:theory}, even when the demonstrations are non-Markovian. In practice, these assumptions do not generally hold, which Equation~(\ref{eq:rfd-population-target}) naturally encapsulates through uncertainty in $p_{\mathcal D}(O^+\mid o^-,P)$. In this case, where $(o^-,P)$ does not uniquely determine $o^+$, $\pi_{\mathrm{RFD}}^\star$ must average the VLA’s distributions over the possible $O^+$ consistent with the information available at $(o^-,P)$. Consequently, $\pi_{\mathrm{RFD}}^\star$ is not in general guaranteed to match $\pi_{\mathrm{VLA}}(\cdot\mid o^+)$; the remaining discrepancy reflects uncertainty over the actual handoff-conditioned VLA distribution that cannot be resolved from $(o^-,P)$ alone.

\subsection{Propose--Resolve}
\label{sec:propose-resolve}
\label{sec:resolution}

At handoff, the observation $o^+$ provides information that was unavailable when the continuation was generated from $o^-$. The distribution $\pi_{\mathrm{VLA}}(\cdot\mid o^+)$ captures how the VLA would react to this newly observed state, but sampling such a reaction requires the very inference delay we seek to avoid. Our second insight is that benefiting from this VLA reaction does not require sampling a new action from it. If multiple continuations $U$ have already been prepared, we can instead evaluate how compatible each is with $\pi_{\mathrm{VLA}}(\cdot\mid o^+)$. Consider $K$ candidate continuations $U^{1:K}$ \textit{proposed} in advance by an asynchronous policy $\pi_{\mathrm{async}}$. Once $o^+$ becomes available, we \textit{resolve} among these candidates by seeking the one with highest likelihood under the handoff-time VLA distribution:
\begin{equation}
k^\star
=
\arg\max_{k\in\{1,\ldots,K\}}
\pi_{\mathrm{VLA}}(U^k\mid o^+).
\label{eq:resolve-ideal}
\end{equation}
This resolution can only begin once $o^+$ has been received. To dispatch the next action without waiting, it must therefore complete within a single control step and be substantially cheaper than generating a new VLA sample. We learn a lightweight conditional density approximation for this purpose. For each training observation $o^+$, we draw $M$ continuations $U_V^{1:M}\sim\pi_{\mathrm{VLA}}(\cdot\mid o^+)$ and estimate their per-coordinate mean $\hat\mu(o^+)$ and standard deviation $\hat\sigma(o^+)$. A lightweight predictor conditioned on $o^+$ learns the corresponding parameters $\mu_\phi(o^+)$ and $\sigma_\phi(o^+)$, defining
\begin{equation}
\hat\pi_\phi(U\mid o^+)
=
\mathcal N\!\left(
U;\mu_\phi(o^+),
\operatorname{diag}\sigma_\phi^2(o^+)
\right).
\label{eq:resolver-gaussian}
\end{equation}
At inference, we use $\hat\pi_\phi$ in place of $\pi_{\mathrm{VLA}}$ in Equation~\ref{eq:resolve-ideal} to rank the prepared continuations. Notably, this replaces the potentially multimodal VLA distribution with a diagonal Gaussian. For action generation, such a unimodal representation is fundamentally problematic in multimodal settings, since averaging across distinct modes can produce actions that correspond to none of them~\citep{chi2023diffusionpolicy}.
For our resolution, however, this failure mode is not necessarily inherited: the Gaussian only ranks already-generated continuations and cannot synthesize an averaged trajectory itself. Intuitively, the burden of representing multimodal action structure therefore remains with the proposer, while the resolver provides a ranking of those proposals informed by $o^+$. Appendix~\ref{app:resolver-analysis} formalizes the relationship between the Gaussian score and $\pi_{\mathrm{VLA}}$, and provides a more grounded analysis, with examples, of when unimodal scoring is and is not informative for multimodal distributions.
Finally, our complete method uses Recursive Flow-Field Distillation to obtain the asynchronous policy $\pi_{\mathrm{RFD}}$ and this policy to generate the proposals for Propose--Resolve.

\section{Experiments}
\label{sec:experiments}

\paragraph{Experimental setup.}
We evaluate a flow-based VLA, SmolVLA~\citep{shukor2025smolvla}, on three RoboMimic tasks~\citep{mandlekar2022robomimic} (ToolHang, Square, and Transport) and on LIBERO~\citep{liu2023libero}. We compare against three approaches to asynchronous execution:
\begin{enumerate*}[label=(\roman*),mode=unboxed]
\item \textsc{rtc}~\citep{black2025rtc} modifies flow generation at inference time to remain compatible with the committed prefix and preceding plan.
\item \textsc{tt-rtc}~\citep{black2025trainingtimertc} instead learns prefix-conditioned continuation generation during training, as analyzed in Section~\ref{sec:theory}.
\item \textsc{paint}~\citep{ho2026paint} retains the frozen policy and uses inference-time inversion and repainting to generate stochastic prefix-compatible continuations.
\end{enumerate*}
Our complete method combines Recursive Flow-Field Distillation (RFD) with Propose--Resolve (PR). Detailed training and evaluation configurations are provided in Appendix~\ref{app:implementation-details}.

\subsection{Distribution Alignment}
\label{sec:recall}
\label{sec:coverage}

We first investigate how the action distributions produced by the asynchronous methods compare with the distribution of the original VLA. In particular, TT-RTC provides an empirical probe of the continuation-target mismatch characterized in Proposition~\ref{prop:continuation-gap}. RFD, in turn, tests whether the future-VLA supervision characterized in Proposition~\ref{prop:rfd-population-target} recovers the VLA action distribution in practice.

% Keep this wrap local to the single handoff-example paragraph.
\begingroup
\setlength{\intextsep}{0pt}
\begin{wrapfigure}{r}{0.5\textwidth}
  \centering
  \includegraphics[width=\linewidth]{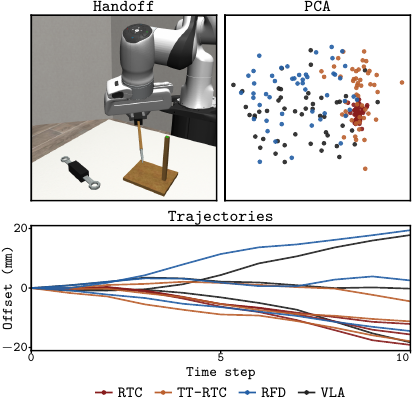}\par
  \vspace{2pt}
  {\fontencoding{T1}\fontfamily{lmtt}\fontsize{8}{9.5}\selectfont
  \setlength{\tabcolsep}{5pt}
  \begin{tabular}{@{}lrr@{}}
    \toprule
    Method & Raw precision (\%) & Raw recall (\%) \\
    \midrule
    \textcolor[HTML]{871C1C}{RTC}    & 99.9 &  0.0 \\
    \textcolor[HTML]{BB602E}{TT-RTC} & 87.1 & 14.1 \\
    \textcolor[HTML]{245FA5}{RFD}    & 60.6 & 86.1 \\
    \bottomrule
    \makebox[0pt][l]{VLA (self-coverage)} & 85.6 & 85.6 \\
  \end{tabular}\par}
  \setlength{\abovecaptionskip}{4pt}
  \caption{\textbf{One ToolHang handoff at $d=10$.}}
  \label{fig:handoff-options}
\end{wrapfigure}

Figure~\ref{fig:handoff-options} exemplifies on a single handoff in ToolHang how we compare the distributions. For the same handoff tuple $(o^-,P,o^+)$, we approximate each asynchronous policy's continuation distribution by sampling many continuations conditioned on $(o^-,P)$, and approximate the VLA distribution by repeatedly sampling from $o^+$.
The trajectory panel shows how the sampled actions evolve through physical space, while the PCA projection makes differences in distributional coverage easier to see. RTC is tightly concentrated, TT-RTC occupies a broader region, and RFD spans a region comparable to the VLA. We quantify this using support recall and precision. Recall measures how much of the VLA's range of possible actions is retained by the asynchronous distribution, while precision measures how much of the asynchronous distribution remains within the range of actions supported by the VLA. Figure~\ref{fig:handoff-options} reports both metrics for this state. Because support is estimated from finite sample banks, even two independent VLA banks achieve only about $85\%$ self-coverage.
\par\WFclear
\endgroup
% Finish the page with this example; the next comparison starts at full width.
\newpage

% Number the handoff example before the quantitative figure on the next page.
\begin{figure}[!t]
  \centering
  \includegraphics{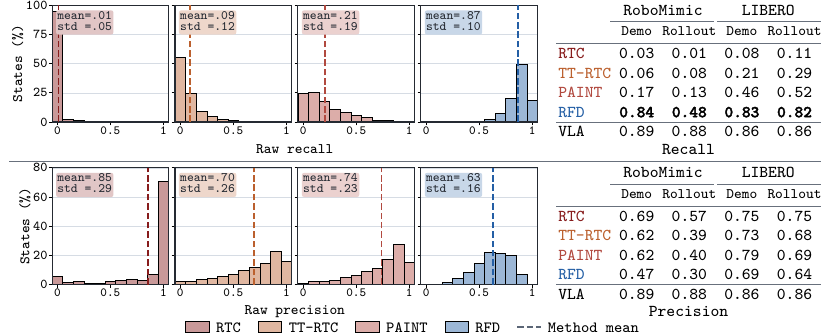}
  \setlength{\abovecaptionskip}{4pt}
  \caption{\textbf{Distribution comparison at $o^+$.}
  For each handoff tuple $(o^-,P,o^+)$ at $d=10$, we draw 128 continuations from each asynchronous method conditioned on $(o^-,P)$ and independent banks of 128 support and 128 query continuations from $\pi_{\mathrm{VLA}}(\cdot\mid o^+)$.
  \textbf{Left:} Per-state precision and recall over $N=2{,}775$ ToolHang demonstration handoffs ($291$ for RTC), shown as histograms.
  \textbf{Right:} Mean precision and recall across RoboMimic and LIBERO, evaluated on both demonstration states and states from VLA rollouts.
  Appendix~\ref{app:coverage} details how precision and recall are computed, extends this analysis across delays $d$, and additionally reports the distribution obtained under the teacher-sample supervision introduced in Section~\ref{sec:flow-field-distillation}.}
  \label{fig:fidelity-histograms}
\end{figure}

The single-state example already shows the pattern we want to measure: RFD reaches approximately the VLA's own self-coverage in recall, whereas TT-RTC recovers only $14\%$ of the VLA support. We now compute precision and recall independently at every handoff state across different ToolHang demonstration trajectories to determine whether this behavior persists beyond a single example. Figure~\ref{fig:fidelity-histograms} (left) shows the resulting distribution of per-state scores.
The aggregate statistics confirm that RTC recovers almost none of the VLA support, while TT-RTC and PAINT broaden the continuation distribution but still miss a substantial part of the VLA's distribution. RFD instead reaches recall close to the VLA self-reference across most states. This supports the prediction motivated by Proposition~\ref{prop:continuation-gap}: replacing demonstration-continuation supervision with supervision from the VLA at handoff recovers substantially more of the action distribution the VLA itself would produce.
Figure~\ref{fig:fidelity-histograms} (right) summarizes the same comparison across RoboMimic and LIBERO, on both demonstration and rollout states. LIBERO shows the same pattern: RFD's recall is close to the VLA self-reference, while the baseline distributions only cover a fraction of that distribution. On RoboMimic rollout states, RFD's recall decreases to $0.48$. Appendix~\ref{app:coverage-generality} gives a per-task comparison, which shows that this drop comes from Square's recall reducing to almost $0$ as $d$ increases.

RFD shows a clear recall--precision asymmetry: its recall remains close to the VLA's self-coverage, while its precision ranges from $35\%$ to $80\%$ of the corresponding VLA self-coverage, depending on the setting. Equation~\ref{eq:rfd-population-target} predicts that RFD matches the handoff-time VLA distribution only when $(o^-,P)$ uniquely determines $o^+$; otherwise, it averages over the VLA distributions associated with possible handoff observations. The observed precision gap is therefore consistent with such averaging before the actual $o^+$ is known, although approximation and finite-sample effects can also contribute. Importantly, high recall alone does not imply better execution: a broader proposal distribution can recover more of the VLA's possible responses while still containing actions that are inappropriate for the particular handoff reached. We therefore next test whether these distributional differences translate into improved closed-loop performance.

\FloatBarrier
\Needspace{6\baselineskip}
\subsection{Real-Time Execution}
\label{sec:latency}

\begin{figure}[!t]
  \centering
  \includegraphics{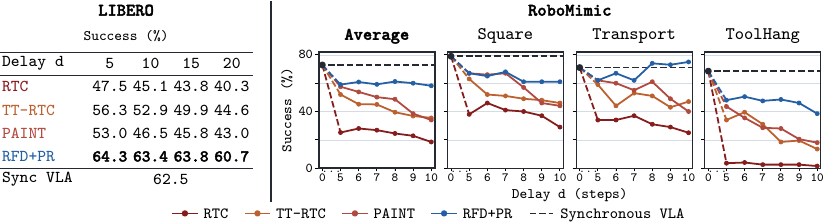}\\[2pt]
  \includegraphics{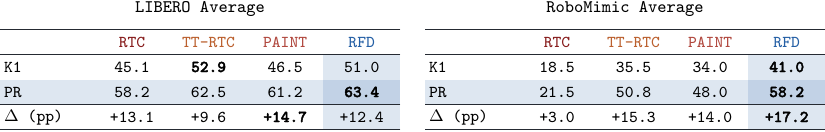}
  \setlength{\abovecaptionskip}{4pt}
  \caption{\textbf{Real-time execution success and ablation.}
  \textbf{Top:} Rollout success on LIBERO and RoboMimic across asynchronous delays, compared with the synchronous VLA reference. Our method uses $K=32$ RFD proposals followed by Propose--Resolve.
  \textbf{Bottom:} Ablation at $d=10$, comparing each asynchronous proposer with a single continuation ($K=1$) and with Propose--Resolve over $K=32$ candidates.}
  \label{fig:rollout-comparison}
  \label{fig:component-contributions}
\end{figure}

Figure~\ref{fig:rollout-comparison} shows closed-loop success across RoboMimic and LIBERO as the asynchronous delay $d$ increases. RFD+PR remains comparatively stable while the asynchronous baselines progressively deteriorate. On RoboMimic, the complete method retains approximately $80\%$ of the original VLA's success, while on LIBERO it matches the original VLA on average. These results show that asynchronous execution can preserve much of the original policy's performance when alternatives are prepared in advance and selected at handoff.

\noindent\textbf{Component attribution.} The ablation in Figure~\ref{fig:component-contributions} (bottom) confirms that the gains do not come from RFD alone. With a single proposal ($K=1$), RFD is strongest on RoboMimic and performs similarly to TT-RTC on LIBERO. After resolving $K=32$ candidates, however, RFD+PR achieves the highest average success on both benchmarks, indicating that the broader proposal distribution becomes most useful when the realized handoff observation can determine which continuation to execute. With SmolVLA, the resolution of the candidates takes approximately $12\,\mathrm{ms}$, so this additional computation step remains compatible with real-time control. Appendix~\ref{app:resolver-implementation} describes the resolver implementation and provides a more detailed timing attribution.

Taken together with Section~\ref{sec:coverage}, these ablations show that the asynchronous methods induce meaningfully different continuation distributions, but the usefulness of those distributions depends strongly on the task. With a single proposal ($K=1$), this difference is already visible in closed-loop success: RTC performs much worse than the broader continuation methods on RoboMimic, whereas all four proposers are considerably closer on LIBERO. Handoff-time selection shows the same contrast: on RoboMimic, PR improves RTC by only $3.0$ percentage points but improves RFD by $17.2$ points, showing that the additional options provided by RFD are particularly valuable for task success. In contrast, on LIBERO, PR improves even RTC by $13.1$ points, and all four proposers end up performing similarly, indicating that the different continuation distributions have much less impact on task success in this setting. Thus, the continuation distribution and task success are not related in a simple one-to-one way: reproducing more of the VLA distribution is important in some settings, while in others substantially different continuation distributions can perform similarly. These results point to a broader tension between renewing behavior from the current observation and preserving coherence with the preceding plan.

\section{Related Work}
\label{sec:related-work}

\noindent\textbf{Generalist Robot Policies.} Generalist robot policies increasingly combine large-scale robot data with pretrained vision-language representations and action heads based on autoregressive prediction, diffusion, or flow matching~\citep{rt1,rt2,openx,octo,kim2024openvla,cogact,black2024pi0,pi05,shukor2025smolvla,gr00t}. Their growing inference cost has motivated work on faster and more efficient execution.

\noindent\textbf{Real-Time VLA Execution.} One direction reduces the cost of the policy query itself, through optimized action heads, parallel prediction, or faster generative decoding~\citep{openvlaoft,flashvla,faster,snapflow}. Our work instead considers the complementary regime in which policy inference remains slower than the desired control rate and must overlap with execution.

\noindent\textbf{Asynchronous Real-Time Execution.} RTC, TT-RTC, and PAINT realize this overlap through different mechanisms for generating prefix-compatible continuations~\citep{black2025rtc,black2025trainingtimertc,ho2026paint}, but all must act from information available before handoff. A growing body of work addresses the resulting staleness by predicting execution-time state or visual features~\citep{tang2025vlash,jiang2026futurertc}, incorporating newly measured state during generation~\citep{park2026pir2}, correcting existing predictions~\citep{a2c2}, or using future observations for supervision or candidate selection~\citep{zhu2026deflect,chen2026dreamchunk}. These approaches address the same stale-conditioning problem through complementary mechanisms. Our focus is instead on characterizing the distributional mismatch induced by continuation learning, aligning the asynchronous proposal distribution with the VLA at handoff, and studying how those proposals can be resolved using the realized handoff observation.

\noindent\textbf{Policy Distillation.} Generative-policy distillation provides another route to efficient control by transferring expensive iterative policies into cheaper generators~\citep{consistencypolicy,onestepdiffusion,dmd,dmd2}. \emph{Recursive Flow-Field Distillation} differs in that teacher and student correspond to different points in the execution timeline, transferring the VLA behavior at handoff into an asynchronous continuation policy conditioned earlier.

\section{Discussion}
\label{sec:discussion}

This work studies asynchronous VLA execution from a distributional perspective. We show that existing continuation objectives generally do not recover the action distribution that the original VLA would produce at handoff, and introduce a new learning objective that instead targets the future VLA distribution. Recursive Flow-Field Distillation provides a practical realization of this objective, and our support analysis provides strong evidence that it learns the intended distribution substantially more faithfully than existing continuation methods. Combined with Propose--Resolve, this enables asynchronous execution with performance close to the original synchronous policy without requiring a new generative VLA pass at handoff. More broadly, our contribution is to frame asynchronous execution as a separation between \emph{which futures should be prepared before handoff} and \emph{how the realized observation should resolve among them afterward}; RFD and PR provide one concrete instantiation of this view.

\noindent\textbf{Limitations.}
Our current resolver has two main limitations. First, although resolution avoids flow generation, it still extracts features from the frozen VLA after $o^+$ arrives. Its latency is therefore partly tied to the underlying VLA architecture; a resolver based on a lightweight state encoder could make the same framework applicable at substantially higher control rates and to slower VLAs. Second, we approximate the handoff-time VLA distribution with a diagonal Gaussian. This deliberately simple model is sufficient for ranking useful proposals in our experiments, but cannot represent the full multimodal structure of the original policy. More expressive conditional density models could provide a stronger resolution rule without changing the proposal mechanism.

\noindent\textbf{Future Work.}
More broadly, our results suggest that reproducing the handoff-time VLA distribution is not always the universally desirable continuation objective. On some tasks, preparing alternatives that resemble a newly queried VLA is important; on others, continuation-based proposals remain equally effective once the reached observation is used for selection. This points to a more general question for asynchronous policies: \emph{what should the policy predict before the future state is known?} Rather than treating continuation of the previous plan or renewal toward the future VLA as universally correct, future work could study when each objective is appropriate and how both can be combined. We view asynchronous execution as a useful setting for studying this broader trade-off between preserving prior intent and renewing behavior from newly available observations.

\newpage
\bibliography{iclr2026_conference}

@inproceedings{black2025rtc,
  author = {Black, Kevin and Galliker, Manuel and Levine, Sergey},
  title = {{Real-Time Execution of Action Chunking Flow Policies}},
  booktitle = {Advances in Neural Information Processing Systems},
  volume = {38, Main Conference},
  year = {2025},
  doi = {10.52202/085713-1122},
  url = {https://proceedings.neurips.cc/paper_files/paper/2025/hash/300ccb2187dedd4edcc07f7e76d8e553-Abstract-Conference.html}
}

@misc{black2025trainingtimertc,
  author = {Kevin Black and Allen Z. Ren and Michael Equi and Sergey Levine},
  title = {{Training-Time Action Conditioning for Efficient Real-Time Chunking}},
  year = {2025},
  eprint = {2512.05964},
  archivePrefix = {arXiv},
  primaryClass = {cs.RO},
  note = {arXiv:2512.05964},
  url = {https://arxiv.org/abs/2512.05964}
}

@misc{ho2026paint,
  author = {Trong-Bao Ho and Quang-Tan Nguyen and Thien-Loc Ha and Gia-Binh Nguyen
    and Viet-Thanh Nguyen and Long Dinh and Minh N. Vu and Duy M. H. Nguyen
    and An Thai Le and Ngo Anh Vien},
  title = {{Start Right, Arrive Right: Asynchronous Execution via Initial Noise Selection}},
  year = {2026},
  eprint = {2606.19774},
  archivePrefix = {arXiv},
  primaryClass = {cs.RO},
  note = {arXiv:2606.19774},
  url = {https://arxiv.org/abs/2606.19774}
}

@misc{kim2024openvla,
  author = {Moo Jin Kim and Karl Pertsch and Siddharth Karamcheti and Ted Xiao and Ashwin Balakrishna and Suraj Nair and Rafael Rafailov and Ethan Foster and Grace Lam and Pannag Sanketi and Quan Vuong and Thomas Kollar and Benjamin Burchfiel and Russ Tedrake and Dorsa Sadigh and Sergey Levine and Percy Liang and Chelsea Finn},
  title = {{OpenVLA: An Open-Source Vision-Language-Action Model}},
  year = {2024},
  eprint = {2406.09246},
  archivePrefix = {arXiv},
  note = {arXiv:2406.09246},
  url = {https://arxiv.org/abs/2406.09246}
}

@misc{black2024pi0,
  author = {Kevin Black and Noah Brown and Danny Driess and Adnan Esmail and Michael Equi and Chelsea Finn and Niccolo Fusai and Lachy Groom and Karol Hausman and Brian Ichter and Szymon Jakubczak and Tim Jones and Liyiming Ke and Sergey Levine and Adrian Li-Bell and Mohith Mothukuri and Suraj Nair and Karl Pertsch and Lucy Xiaoyang Shi and James Tanner and Quan Vuong and Anna Walling and Haohuan Wang and Ury Zhilinsky},
  title = {{$\pi_0$: A Vision-Language-Action Flow Model for General Robot Control}},
  year = {2024},
  eprint = {2410.24164},
  archivePrefix = {arXiv},
  note = {arXiv:2410.24164},
  url = {https://arxiv.org/abs/2410.24164}
}

@misc{shukor2025smolvla,
  author = {Mustafa Shukor and Dana Aubakirova and Francesco Capuano and Pepijn Kooijmans and Steven Palma and Adil Zouitine and Michel Aractingi and Caroline Pascal and Martino Russi and Andres Marafioti and Simon Alibert and Matthieu Cord and Thomas Wolf and Remi Cadene},
  title = {{SmolVLA: A Vision-Language-Action Model for Affordable and Efficient Robotics}},
  year = {2025},
  eprint = {2506.01844},
  archivePrefix = {arXiv},
  note = {arXiv:2506.01844},
  url = {https://arxiv.org/abs/2506.01844}
}

@misc{tang2025vlash,
  author = {Jiaming Tang and Yufei Sun and Yilong Zhao and Shang Yang and Yujun Lin and Zhuoyang Zhang and James Hou and Yao Lu and Zhijian Liu and Song Han},
  title = {{VLASH: Real-Time VLAs via Future-State-Aware Asynchronous Inference}},
  year = {2025},
  eprint = {2512.01031},
  archivePrefix = {arXiv},
  note = {arXiv:2512.01031},
  url = {https://arxiv.org/abs/2512.01031}
}

@misc{park2026pir2,
  author = {Sungjae Park and Shubham Tulsiani},
  title = {{$\pi\mathbf{R}^2$: Reactive Real-time Flow Policies}},
  year = {2026},
  eprint = {2607.26055},
  archivePrefix = {arXiv},
  note = {arXiv:2607.26055},
  url = {https://arxiv.org/abs/2607.26055}
}

@misc{jiang2026futurertc,
  author = {Hai Jiang and Yixian Zou and Binbin Liang and Boqian Liu and Fanman Meng and Shuaicheng Liu},
  title = {{FutureRTC: Real-Time Robot Execution with Anticipatory-Conditioned Action Chunking}},
  year = {2026},
  eprint = {2607.24008},
  archivePrefix = {arXiv},
  note = {arXiv:2607.24008},
  url = {https://arxiv.org/abs/2607.24008}
}

@inproceedings{zhao2023act,
  author = {Tony Z. Zhao and Vikash Kumar and Sergey Levine and Chelsea Finn},
  title = {{Learning Fine-Grained Bimanual Manipulation with Low-Cost Hardware}},
  booktitle = {Proceedings of Robotics: Science and Systems},
  year = {2023},
  doi = {10.15607/RSS.2023.XIX.016},
  url = {https://roboticsproceedings.org/rss19/p016.html}
}

@inproceedings{chi2023diffusionpolicy,
  author = {Cheng Chi and Siyuan Feng and Yilun Du and Zhenjia Xu and Eric Cousineau and Benjamin CM Burchfiel and Shuran Song},
  title = {{Diffusion Policy: Visuomotor Policy Learning via Action Diffusion}},
  booktitle = {Proceedings of Robotics: Science and Systems},
  year = {2023},
  doi = {10.15607/RSS.2023.XIX.026},
  url = {https://roboticsproceedings.org/rss19/p026.html}
}

@misc{zhu2026deflect,
  author = {Yixiang Zhu and Yonghao Chen and Zijie Yang and Yusong Hu and Xinyu Chen},
  title = {{DEFLECT: Temporal Counterfactual Preference Learning for Delay-Robust Asynchronous VLAs}},
  year = {2026},
  eprint = {2605.19294},
  archivePrefix = {arXiv},
  primaryClass = {cs.RO},
  note = {arXiv:2605.19294},
  url = {https://arxiv.org/abs/2605.19294}
}

@misc{chen2026dreamchunk,
  author = {Wenxi Chen and Kaidi Zhang and Chi Lin and Zhiyuan Zhang and Yu She and Yuejiang Liu and Raymond A. Yeh and Shaoshuai Mou and Yan Gu},
  title = {{DREAM-Chunk: Reactive Action Chunking with Latent World Model}},
  year = {2026},
  eprint = {2606.18589},
  archivePrefix = {arXiv},
  primaryClass = {cs.RO},
  note = {arXiv:2606.18589},
  url = {https://arxiv.org/abs/2606.18589}
}

@misc{rt1,
      title={{RT-1: Robotics Transformer for Real-World Control at Scale}}, 
      author={Anthony Brohan and Noah Brown and Justice Carbajal and Yevgen Chebotar and Joseph Dabis and Chelsea Finn and Keerthana Gopalakrishnan and Karol Hausman and Alex Herzog and Jasmine Hsu and Julian Ibarz and Brian Ichter and Alex Irpan and Tomas Jackson and Sally Jesmonth and Nikhil J Joshi and Ryan Julian and Dmitry Kalashnikov and Yuheng Kuang and Isabel Leal and Kuang-Huei Lee and Sergey Levine and Yao Lu and Utsav Malla and Deeksha Manjunath and Igor Mordatch and Ofir Nachum and Carolina Parada and Jodilyn Peralta and Emily Perez and Karl Pertsch and Jornell Quiambao and Kanishka Rao and Michael Ryoo and Grecia Salazar and Pannag Sanketi and Kevin Sayed and Jaspiar Singh and Sumedh Sontakke and Austin Stone and Clayton Tan and Huong Tran and Vincent Vanhoucke and Steve Vega and Quan Vuong and Fei Xia and Ted Xiao and Peng Xu and Sichun Xu and Tianhe Yu and Brianna Zitkovich},
      year={2023},
      eprint={2212.06817},
      archivePrefix={arXiv},
      primaryClass={cs.RO},
      note={arXiv:2212.06817},
      url={https://arxiv.org/abs/2212.06817}, 
}

@misc{rt2,
      title={{RT-2: Vision-Language-Action Models Transfer Web Knowledge to Robotic Control}}, 
      author={Anthony Brohan and Noah Brown and Justice Carbajal and Yevgen Chebotar and Xi Chen and Krzysztof Choromanski and Tianli Ding and Danny Driess and Avinava Dubey and Chelsea Finn and Pete Florence and Chuyuan Fu and Montse Gonzalez Arenas and Keerthana Gopalakrishnan and Kehang Han and Karol Hausman and Alexander Herzog and Jasmine Hsu and Brian Ichter and Alex Irpan and Nikhil Joshi and Ryan Julian and Dmitry Kalashnikov and Yuheng Kuang and Isabel Leal and Lisa Lee and Tsang-Wei Edward Lee and Sergey Levine and Yao Lu and Henryk Michalewski and Igor Mordatch and Karl Pertsch and Kanishka Rao and Krista Reymann and Michael Ryoo and Grecia Salazar and Pannag Sanketi and Pierre Sermanet and Jaspiar Singh and Anikait Singh and Radu Soricut and Huong Tran and Vincent Vanhoucke and Quan Vuong and Ayzaan Wahid and Stefan Welker and Paul Wohlhart and Jialin Wu and Fei Xia and Ted Xiao and Peng Xu and Sichun Xu and Tianhe Yu and Brianna Zitkovich},
      year={2023},
      eprint={2307.15818},
      archivePrefix={arXiv},
      primaryClass={cs.RO},
      note={arXiv:2307.15818},
      url={https://arxiv.org/abs/2307.15818}, 
}

@misc{openx,
      title={{Open X-Embodiment: Robotic Learning Datasets and RT-X Models}}, 
      author={{Embodiment Collaboration} and Abby O'Neill and Abdul Rehman and Abhinav Gupta and Abhiram Maddukuri and Abhishek Gupta and Abhishek Padalkar and Abraham Lee and Acorn Pooley and Agrim Gupta and Ajay Mandlekar and Ajinkya Jain and Albert Tung and Alex Bewley and Alex Herzog and Alex Irpan and Alexander Khazatsky and Anant Rai and Anchit Gupta and Andrew Wang and Andrey Kolobov and Anikait Singh and Animesh Garg and Aniruddha Kembhavi and Annie Xie and Anthony Brohan and Antonin Raffin and Archit Sharma and Arefeh Yavary and Arhan Jain and Ashwin Balakrishna and Ayzaan Wahid and Ben Burgess-Limerick and Beomjoon Kim and Bernhard Schölkopf and Blake Wulfe and Brian Ichter and Cewu Lu and Charles Xu and Charlotte Le and Chelsea Finn and Chen Wang and Chenfeng Xu and Cheng Chi and Chenguang Huang and Christine Chan and Christopher Agia and Chuer Pan and Chuyuan Fu and Coline Devin and Danfei Xu and Daniel Morton and Danny Driess and Daphne Chen and Deepak Pathak and Dhruv Shah and Dieter Büchler and Dinesh Jayaraman and Dmitry Kalashnikov and Dorsa Sadigh and Edward Johns and Ethan Foster and Fangchen Liu and Federico Ceola and Fei Xia and Feiyu Zhao and Felipe Vieira Frujeri and Freek Stulp and Gaoyue Zhou and Gaurav S. Sukhatme and Gautam Salhotra and Ge Yan and Gilbert Feng and Giulio Schiavi and Glen Berseth and Gregory Kahn and Guangwen Yang and Guanzhi Wang and Hao Su and Hao-Shu Fang and Haochen Shi and Henghui Bao and Heni Ben Amor and Henrik I Christensen and Hiroki Furuta and Homanga Bharadhwaj and Homer Walke and Hongjie Fang and Huy Ha and Igor Mordatch and Ilija Radosavovic and Isabel Leal and Jacky Liang and Jad Abou-Chakra and Jaehyung Kim and Jaimyn Drake and Jan Peters and Jan Schneider and Jasmine Hsu and Jay Vakil and Jeannette Bohg and Jeffrey Bingham and Jeffrey Wu and Jensen Gao and Jiaheng Hu and Jiajun Wu and Jialin Wu and Jiankai Sun and Jianlan Luo and Jiayuan Gu and Jie Tan and Jihoon Oh and Jimmy Wu and Jingpei Lu and Jingyun Yang and Jitendra Malik and João Silvério and Joey Hejna and Jonathan Booher and Jonathan Tompson and Jonathan Yang and Jordi Salvador and Joseph J. Lim and Junhyek Han and Kaiyuan Wang and Kanishka Rao and Karl Pertsch and Karol Hausman and Keegan Go and Keerthana Gopalakrishnan and Ken Goldberg and Kendra Byrne and Kenneth Oslund and Kento Kawaharazuka and Kevin Black and Kevin Lin and Kevin Zhang and Kiana Ehsani and Kiran Lekkala and Kirsty Ellis and Krishan Rana and Krishnan Srinivasan and Kuan Fang and Kunal Pratap Singh and Kuo-Hao Zeng and Kyle Hatch and Kyle Hsu and Laurent Itti and Lawrence Yunliang Chen and Lerrel Pinto and Li Fei-Fei and Liam Tan and Linxi "Jim" Fan and Lionel Ott and Lisa Lee and Luca Weihs and Magnum Chen and Marion Lepert and Marius Memmel and Masayoshi Tomizuka and Masha Itkina and Mateo Guaman Castro and Max Spero and Maximilian Du and Michael Ahn and Michael C. Yip and Mingtong Zhang and Mingyu Ding and Minho Heo and Mohan Kumar Srirama and Mohit Sharma and Moo Jin Kim and Muhammad Zubair Irshad and Naoaki Kanazawa and Nicklas Hansen and Nicolas Heess and Nikhil J Joshi and Niko Suenderhauf and Ning Liu and Norman Di Palo and Nur Muhammad Mahi Shafiullah and Oier Mees and Oliver Kroemer and Osbert Bastani and Pannag R Sanketi and Patrick "Tree" Miller and Patrick Yin and Paul Wohlhart and Peng Xu and Peter David Fagan and Peter Mitrano and Pierre Sermanet and Pieter Abbeel and Priya Sundaresan and Qiuyu Chen and Quan Vuong and Rafael Rafailov and Ran Tian and Ria Doshi and Roberto Martín-Martín and Rohan Baijal and Rosario Scalise and Rose Hendrix and Roy Lin and Runjia Qian and Ruohan Zhang and Russell Mendonca and Rutav Shah and Ryan Hoque and Ryan Julian and Samuel Bustamante and Sean Kirmani and Sergey Levine and Shan Lin and Sherry Moore and Shikhar Bahl and Shivin Dass and Shubham Sonawani and Shubham Tulsiani and Shuran Song and Sichun Xu and Siddhant Haldar and Siddharth Karamcheti and Simeon Adebola and Simon Guist and Soroush Nasiriany and Stefan Schaal and Stefan Welker and Stephen Tian and Subramanian Ramamoorthy and Sudeep Dasari and Suneel Belkhale and Sungjae Park and Suraj Nair and Suvir Mirchandani and Takayuki Osa and Tanmay Gupta and Tatsuya Harada and Tatsuya Matsushima and Ted Xiao and Thomas Kollar and Tianhe Yu and Tianli Ding and Todor Davchev and Tony Z. Zhao and Travis Armstrong and Trevor Darrell and Trinity Chung and Vidhi Jain and Vikash Kumar and Vincent Vanhoucke and Vitor Guizilini and Wei Zhan and Wenxuan Zhou and Wolfram Burgard and Xi Chen and Xiangyu Chen and Xiaolong Wang and Xinghao Zhu and Xinyang Geng and Xiyuan Liu and Xu Liangwei and Xuanlin Li and Yansong Pang and Yao Lu and Yecheng Jason Ma and Yejin Kim and Yevgen Chebotar and Yifan Zhou and Yifeng Zhu and Yilin Wu and Ying Xu and Yixuan Wang and Yonatan Bisk and Yongqiang Dou and Yoonyoung Cho and Youngwoon Lee and Yuchen Cui and Yue Cao and Yueh-Hua Wu and Yujin Tang and Yuke Zhu and Yunchu Zhang and Yunfan Jiang and Yunshuang Li and Yunzhu Li and Yusuke Iwasawa and Yutaka Matsuo and Zehan Ma and Zhuo Xu and Zichen Jeff Cui and Zichen Zhang and Zipeng Fu and Zipeng Lin},
      year={2025},
      eprint={2310.08864},
      archivePrefix={arXiv},
      primaryClass={cs.RO},
      note={arXiv:2310.08864},
      url={https://arxiv.org/abs/2310.08864}, 
}

@misc{octo,
      title={{Octo: An Open-Source Generalist Robot Policy}}, 
      author={{Octo Model Team} and Dibya Ghosh and Homer Walke and Karl Pertsch and Kevin Black and Oier Mees and Sudeep Dasari and Joey Hejna and Tobias Kreiman and Charles Xu and Jianlan Luo and You Liang Tan and Lawrence Yunliang Chen and Pannag Sanketi and Quan Vuong and Ted Xiao and Dorsa Sadigh and Chelsea Finn and Sergey Levine},
      year={2024},
      eprint={2405.12213},
      archivePrefix={arXiv},
      primaryClass={cs.RO},
      note={arXiv:2405.12213},
      url={https://arxiv.org/abs/2405.12213}, 
}

@misc{cogact,
      title={{CogACT: A Foundational Vision-Language-Action Model for Synergizing Cognition and Action in Robotic Manipulation}}, 
      author={Qixiu Li and Yaobo Liang and Zeyu Wang and Lin Luo and Xi Chen and Mozheng Liao and Fangyun Wei and Yu Deng and Sicheng Xu and Yizhong Zhang and Xiaofan Wang and Bei Liu and Jianlong Fu and Jianmin Bao and Dong Chen and Yuanchun Shi and Jiaolong Yang and Baining Guo},
      year={2024},
      eprint={2411.19650},
      archivePrefix={arXiv},
      primaryClass={cs.RO},
      note={arXiv:2411.19650},
      url={https://arxiv.org/abs/2411.19650}, 
}

@misc{pi05,
      title={{$\pi_{0.5}$: a Vision-Language-Action Model with Open-World Generalization}}, 
      author={{Physical Intelligence} and Kevin Black and Noah Brown and James Darpinian and Karan Dhabalia and Danny Driess and Adnan Esmail and Michael Equi and Chelsea Finn and Niccolo Fusai and Manuel Y. Galliker and Dibya Ghosh and Lachy Groom and Karol Hausman and Brian Ichter and Szymon Jakubczak and Tim Jones and Liyiming Ke and Devin LeBlanc and Sergey Levine and Adrian Li-Bell and Mohith Mothukuri and Suraj Nair and Karl Pertsch and Allen Z. Ren and Lucy Xiaoyang Shi and Laura Smith and Jost Tobias Springenberg and Kyle Stachowicz and James Tanner and Quan Vuong and Homer Walke and Anna Walling and Haohuan Wang and Lili Yu and Ury Zhilinsky},
      year={2025},
      eprint={2504.16054},
      archivePrefix={arXiv},
      primaryClass={cs.LG},
      note={arXiv:2504.16054},
      url={https://arxiv.org/abs/2504.16054}, 
}

@misc{gr00t,
      title={{GR00T N1: An Open Foundation Model for Generalist Humanoid Robots}}, 
      author={{NVIDIA} and Johan Bjorck and Fernando Castañeda and Nikita Cherniadev and Xingye Da and Runyu Ding and Linxi "Jim" Fan and Yu Fang and Dieter Fox and Fengyuan Hu and Spencer Huang and Joel Jang and Zhenyu Jiang and Jan Kautz and Kaushil Kundalia and Lawrence Lao and Zhiqi Li and Zongyu Lin and Kevin Lin and Guilin Liu and Edith Llontop and Loic Magne and Ajay Mandlekar and Avnish Narayan and Soroush Nasiriany and Scott Reed and You Liang Tan and Guanzhi Wang and Zu Wang and Jing Wang and Qi Wang and Jiannan Xiang and Yuqi Xie and Yinzhen Xu and Zhenjia Xu and Seonghyeon Ye and Zhiding Yu and Ao Zhang and Hao Zhang and Yizhou Zhao and Ruijie Zheng and Yuke Zhu},
      year={2025},
      eprint={2503.14734},
      archivePrefix={arXiv},
      primaryClass={cs.RO},
      note={arXiv:2503.14734},
      url={https://arxiv.org/abs/2503.14734}, 
}

@misc{openvlaoft,
      title={{Fine-Tuning Vision-Language-Action Models: Optimizing Speed and Success}}, 
      author={Moo Jin Kim and Chelsea Finn and Percy Liang},
      year={2025},
      eprint={2502.19645},
      archivePrefix={arXiv},
      primaryClass={cs.RO},
      note={arXiv:2502.19645},
      url={https://arxiv.org/abs/2502.19645}, 
}

@misc{flashvla,
      title={{FlashVLA: Streaming Action Decoding for Fast and Asynchronous VLA Inference}}, 
      author={Zekai Li and Jiaming Tang and Zhijian Liu},
      year={2026},
      eprint={2608.27384},
      archivePrefix={arXiv},
      primaryClass={cs.RO},
      note={arXiv:2608.27384},
      url={https://arxiv.org/abs/2608.27384}, 
}

@misc{faster,
      title={{FASTER: Rethinking Real-Time Flow VLAs}}, 
      author={Yuxiang Lu and Zhe Liu and Xianzhe Fan and Zhenya Yang and Jinghua Hou and Junyi Li and Kaixin Ding and Hengshuang Zhao},
      year={2026},
      eprint={2603.19199},
      archivePrefix={arXiv},
      primaryClass={cs.RO},
      note={arXiv:2603.19199},
      url={https://arxiv.org/abs/2603.19199}, 
}

@misc{snapflow,
      title={{SnapFlow: One-Step Action Generation for Flow-Matching VLAs via Progressive Self-Distillation}},
      author={Wuyang Luan and Junhui Li and Weiguang Zhao and Wenjian Zhang and Tieru Wu and Rui Ma},
      year={2026},
      eprint={2604.05656},
      archivePrefix={arXiv},
      primaryClass={cs.CV},
      note={arXiv:2604.05656},
      url={https://arxiv.org/abs/2604.05656},
}

@misc{a2c2,
      title={{Leave No Observation Behind: Real-time Correction for VLA Action Chunks}}, 
      author={Kohei Sendai and Maxime Alvarez and Tatsuya Matsushima and Yutaka Matsuo and Yusuke Iwasawa},
      year={2025},
      eprint={2509.23224},
      archivePrefix={arXiv},
      primaryClass={cs.RO},
      note={arXiv:2509.23224},
      url={https://arxiv.org/abs/2509.23224}, 
}

@misc{consistencypolicy,
      title={{Consistency Policy: Accelerated Visuomotor Policies via Consistency Distillation}}, 
      author={Aaditya Prasad and Kevin Lin and Jimmy Wu and Linqi Zhou and Jeannette Bohg},
      year={2024},
      eprint={2405.07503},
      archivePrefix={arXiv},
      primaryClass={cs.RO},
      note={arXiv:2405.07503},
      url={https://arxiv.org/abs/2405.07503}, 
}

@misc{onestepdiffusion,
      title={{One-Step Diffusion Policy: Fast Visuomotor Policies via Diffusion Distillation}}, 
      author={Zhendong Wang and Zhaoshuo Li and Ajay Mandlekar and Zhenjia Xu and Jiaojiao Fan and Yashraj Narang and Linxi Fan and Yuke Zhu and Yogesh Balaji and Mingyuan Zhou and Ming-Yu Liu and Yu Zeng},
      year={2024},
      eprint={2410.21257},
      archivePrefix={arXiv},
      primaryClass={cs.RO},
      note={arXiv:2410.21257},
      url={https://arxiv.org/abs/2410.21257}, 
}

@misc{dmd,
      title={{One-step Diffusion with Distribution Matching Distillation}}, 
      author={Tianwei Yin and Michaël Gharbi and Richard Zhang and Eli Shechtman and Fredo Durand and William T. Freeman and Taesung Park},
      year={2024},
      eprint={2311.18828},
      archivePrefix={arXiv},
      primaryClass={cs.CV},
      note={arXiv:2311.18828},
      url={https://arxiv.org/abs/2311.18828}, 
}

@misc{dmd2,
      title={{Improved Distribution Matching Distillation for Fast Image Synthesis}}, 
      author={Tianwei Yin and Michaël Gharbi and Taesung Park and Richard Zhang and Eli Shechtman and Fredo Durand and William T. Freeman},
      year={2024},
      eprint={2405.14867},
      archivePrefix={arXiv},
      primaryClass={cs.CV},
      note={arXiv:2405.14867},
      url={https://arxiv.org/abs/2405.14867}, 
}

@misc{lazzati2026chunking,
      title={{Why Does Action Chunking Improve Behavioral Cloning Performance in Robotic Control?}}, 
      author={Filippo Lazzati and Kyle Stachowicz and William Chen and Alberto Maria Metelli and Andrew Wagenmaker and Sergey Levine},
      year={2026},
      eprint={2608.02547},
      archivePrefix={arXiv},
      primaryClass={cs.RO},
      note={arXiv:2608.02547},
      url={https://arxiv.org/abs/2608.02547}, 
}

@misc{zeng2026openloop,
      title={{Revisiting Open-Loop Execution in Robotics: Toward Reactive, Higher-Performing Policies}}, 
      author={Michael Zeng and Abhinav Agarwal and Ajay Bati and Brian Lee and Siddharth Ancha and Russ Tedrake},
      year={2026},
      eprint={2608.15938},
      archivePrefix={arXiv},
      primaryClass={cs.RO},
      note={arXiv:2608.15938},
      url={https://arxiv.org/abs/2608.15938}, 
}

@inproceedings{lipman2023flowmatching,
  author = {Yaron Lipman and Ricky T. Q. Chen and Heli Ben-Hamu and Maximilian Nickel and Matt Le},
  title = {{Flow Matching for Generative Modeling}},
  booktitle = {The Eleventh International Conference on Learning Representations},
  year = {2023},
  url = {https://openreview.net/forum?id=PqvMRDCJT9t}
}

@book{cover2006elements,
  author = {Thomas M. Cover and Joy A. Thomas},
  title = {{Elements of Information Theory}},
  edition = {2},
  publisher = {John Wiley \& Sons},
  year = {2006},
  doi = {10.1002/047174882X},
  isbn = {9780471241959},
  url = {https://onlinelibrary.wiley.com/doi/book/10.1002/047174882X}
}

@inproceedings{mandlekar2022robomimic,
  author = {Ajay Mandlekar and Danfei Xu and Josiah Wong and Soroush Nasiriany and Chen Wang and Rohun Kulkarni and Li Fei-Fei and Silvio Savarese and Yuke Zhu and Roberto Mart\'{i}n-Mart\'{i}n},
  title = {{What Matters in Learning from Offline Human Demonstrations for Robot Manipulation}},
  booktitle = {Proceedings of the 5th Conference on Robot Learning},
  pages = {1678--1690},
  year = {2022},
  volume = {164},
  series = {Proceedings of Machine Learning Research},
  publisher = {PMLR},
  url = {https://proceedings.mlr.press/v164/mandlekar22a.html}
}

@inproceedings{liu2023libero,
  author = {Bo Liu and Yifeng Zhu and Chongkai Gao and Yihao Feng and Qiang Liu and Yuke Zhu and Peter Stone},
  title = {{LIBERO: Benchmarking Knowledge Transfer for Lifelong Robot Learning}},
  booktitle = {Advances in Neural Information Processing Systems},
  pages = {44776--44791},
  year = {2023},
  volume = {36},
  publisher = {Curran Associates, Inc.},
  doi = {10.52202/075280-1939},
  url = {https://papers.nips.cc/paper_files/paper/2023/hash/8c3c666820ea055a77726d66fc7d447f-Abstract-Datasets_and_Benchmarks.html}
}

@inproceedings{minka2001expectation,
  author = {Thomas P. Minka},
  title = {{Expectation Propagation for Approximate Bayesian Inference}},
  booktitle = {Proceedings of the Seventeenth Conference on Uncertainty in Artificial Intelligence},
  pages = {362--369},
  year = {2001},
  url = {https://tminka.github.io/papers/ep/}
}

@inproceedings{kynkaanniemi2019precision,
  author = {Kynk{\"a}{\"a}nniemi, Tuomas and Karras, Tero and Laine, Samuli and Lehtinen, Jaakko and Aila, Timo},
  title = {{Improved Precision and Recall Metric for Assessing Generative Models}},
  booktitle = {Advances in Neural Information Processing Systems},
  volume = {32},
  year = {2019},
  url = {https://proceedings.neurips.cc/paper/2019/hash/0234c510bc6d908b28c70ff313743079-Abstract.html}
}
\bibliographystyle{iclr2027_conference}

\clearpage
\appendix
\section{Continuation Learning and VLA Replanning}
\label{app:target-gap}

\subsection{Proof of Proposition~\ref{prop:continuation-gap}}
\label{app:proof-continuation-gap}

\paragraph{Proof.}
All expectations and conditional distributions are taken under
$\mathcal D$. The conditional-KL representation of mutual
information gives~\citep[Ch.~2]{cover2006elements}
\begin{equation}
I_{\mathcal D}(U;O^-,P\mid O^+)
=
\mathbb E_{O^-,P,O^+}\!\left[
D_{\mathrm{KL}}\!\left(
p_{\mathcal D}(U\mid O^-,P,O^+)
\,\Vert\,
p_{\mathcal D}(U\mid O^+)
\right)
\right].
\end{equation}
Because $O^+=f(O^-,P)$, additionally conditioning on $O^+$
does not change $p_{\mathcal D}(U\mid O^-,P)$, and the outer
expectation over $O^+$ is redundant. Substitution therefore
yields Eq.~\ref{eq:continuation-gap}. By nonnegativity of KL,
this expectation is zero exactly when its two conditional
distributions agree almost everywhere. Conditional mutual
information is zero exactly when
$U\perp(O^-,P)\mid O^+$, establishing
Eq.~\ref{eq:continuation-equivalence}. \hfill$\square$

\paragraph{A minimal deterministic counterexample.}
Consider fully observed scalar dynamics
$O_{k+1}=O_k+a_k$, with a one-action prefix and a one-action
continuation ($d=s=1$). The dataset contains two equally
likely trajectories:
\begin{equation}
-1 \xrightarrow{P=+1} 0 \xrightarrow{U=+1} +1,
\qquad
+1 \xrightarrow{P=-1} 0 \xrightarrow{U=-1} -1.
\end{equation}
In each demonstration, the demonstrator repeats the action
used to reach the origin. Both trajectories therefore have
the same handoff observation, but the prefix reveals which
continuation follows. Specifically,
\begin{equation}
p_{\mathcal D}(U\mid O^+=0)
=
\tfrac12\delta_{+1}+\tfrac12\delta_{-1},
\qquad
p_{\mathcal D}(U\mid O^-=-1,P=+1)=\delta_{+1},
\end{equation}
where $\delta_a$ denotes a point mass at $a$; the other
earlier context similarly gives $\delta_{-1}$. Each
prefix-conditioned distribution has KL divergence $\log 2$
from the handoff-conditioned mixture, so the expected gap
is $\log 2$ nats. No uncertainty about the physical state
is involved: the difference arises because demonstration
behavior depends on how that state was reached. Under exact
population learning, TT-RTC retains this dependence, whereas
the observation-conditioned VLA represents both continuations.

\subsection{Beyond Deterministic Handoffs}
\label{app:stochastic-handoffs}

Proposition~\ref{prop:continuation-gap} deliberately removes
uncertainty about the reached observation to isolate the
effect of history-dependent demonstration behavior. We now
drop this restriction while continuing to compare the
population targets in Eq.~\ref{eq:two-prediction-problems}.

Uncertainty in $p_{\mathcal D}(O^+\mid O^-,P)$ can arise
from stochastic dynamics or noisy observations. Partial
observability can produce the same effect even with
deterministic physical dynamics: different hidden states
consistent with the earlier observation can lead to
different handoff observations after the same prefix.
The following analysis accommodates these cases directly
through the joint distribution of the observed variables.

\paragraph{Comparison using the information available before handoff.}
The demonstration-continuation target averages over possible
handoff observations. For comparison, define the corresponding
average of the handoff-conditioned target:
\begin{equation}
\begin{aligned}
p_{\mathcal D}(U\mid O^-,P)
&=
\mathbb E_{O^+\mid O^-,P}\!\left[
p_{\mathcal D}(U\mid O^-,P,O^+)
\right],\\
\bar p_{\mathcal D}(U\mid O^-,P)
&:=
\mathbb E_{O^+\mid O^-,P}\!\left[
p_{\mathcal D}(U\mid O^+)
\right].
\end{aligned}
\label{eq:stochastic-continuation-targets}
\end{equation}
Both expressions use the same distribution over reached
observations. They differ in whether the continuation
distribution retains the earlier observation and prefix
after the handoff observation is specified. Consequently,
$U\perp(O^-,P)\mid O^+$ remains sufficient for these two
averaged targets to coincide.

Applying joint convexity of KL to
Eq.~\ref{eq:stochastic-continuation-targets}, then averaging
over $(O^-,P)$, gives
\begin{equation}
\begin{aligned}
&\mathbb E_{O^-,P}\!\left[
D_{\mathrm{KL}}\!\left(
p_{\mathcal D}(U\mid O^-,P)
\,\Vert\,
\bar p_{\mathcal D}(U\mid O^-,P)
\right)
\right]\\
&\quad\le
\mathbb E_{O^-,P,O^+}\!\left[
D_{\mathrm{KL}}\!\left(
p_{\mathcal D}(U\mid O^-,P,O^+)
\,\Vert\,
p_{\mathcal D}(U\mid O^+)
\right)
\right]\\
&\quad=
I_{\mathcal D}(U;O^-,P\mid O^+).
\end{aligned}
\label{eq:stochastic-continuation-bound}
\end{equation}
When $O^+=f(O^-,P)$, each mixture reduces to a single
conditional distribution and the bound becomes the equality
in Proposition~\ref{prop:continuation-gap}. Otherwise,
averaging can cancel differences between the conditional
distributions. The conditional-independence condition is
therefore sufficient, but no longer necessary, for equality
of the averaged targets. Stochasticity should not be
interpreted as necessarily increasing the KL mismatch.

\paragraph{Comparison at the realized handoff.}
Equality to the averaged reference
$\bar p_{\mathcal D}(U\mid O^-,P)$ does not imply equality to
$p_{\mathcal D}(U\mid o^+)$ at the observation actually reached.
Even when the demonstrations satisfy the conditional-independence
condition, different possible handoff observations may induce
different continuation distributions. A single distribution
conditioned only on $(o^-,P)$ cannot equal all of them unless
they coincide across the possible handoff observations.

There are thus two distinct information questions: whether
the earlier context adds information about demonstrated
behavior once $O^+$ is known, and whether observing $O^+$
adds information about the relevant handoff-conditioned target.
Proposition~\ref{prop:continuation-gap} isolates the first
by eliminating the second. Under partial observability,
history may also reveal hidden physical state, so even an
expert that is Markovian in the full state need not satisfy
the same property with respect to the observation alone.

\section{Population Analysis of Recursive Flow-Field Distillation}
\label{app:teacher-supervision}

\paragraph{Proof of Proposition~\ref{prop:rfd-population-target}.}
Fix $(o^-,P)$, let $d$ be the prefix length, and set $m=H-d$.
All expectations below are conditional on this fixed context.
Draw $O^+\sim p_{\mathcal D}(\cdot\mid o^-,P)$, then a full
$H$-step teacher chunk $B_V$ conditioned only on $O^+$, and
independent Gaussian noise $\epsilon$.
With $Z_\tau=\tau\epsilon+(1-\tau)B_V$, write
$X_\tau=Z_{\tau,0:m}$.
We assume finite second moments, independent flow-time sampling
with positive density on $(0,1)$, unrestricted population
optimization, and sufficient regularity for a unique probability
flow with well-defined endpoint limits. The exact-teacher
assumption means that
\begin{equation}
v_{\mathrm{VLA}}(z,\tau\mid o^+)
=
\mathbb E\!\left[
\epsilon-B_V
\mid Z_\tau=z,\tau,O^+=o^+
\right].
\label{eq:exact-teacher-field}
\end{equation}

Denote the student's continuation field by
$w(x,\tau):=v_{\mathrm{RFD}}([P;x],\tau\mid o^-)_{d:H}$
and its teacher target by
$T_\tau:=v_{\mathrm{VLA}}(Z_\tau,\tau\mid O^+)_{0:m}$.
The population MSE minimizer is the conditional mean of its
target. Applying Eq.~\ref{eq:exact-teacher-field} and the tower
property gives
\begin{equation}
\begin{aligned}
w^\star(x,\tau)
&=\mathbb E[T_\tau\mid X_\tau=x,\tau]\\
&=\mathbb E[(\epsilon-B_V)_{0:m}\mid X_\tau=x,\tau].
\end{aligned}
\label{eq:rfd-optimal-field}
\end{equation}
The second equality uses the conditional independence of the
teacher sample and noise from the earlier context given $O^+$.
It averages over both the unknown handoff observation and the
teacher's omitted last $d$ coordinates.

Since $(\epsilon-B_V)_{0:m}=\partial_\tau X_\tau$,
Eq.~\ref{eq:rfd-optimal-field} is the marginal flow-matching
field for the probability path of $X_\tau$
\citep{lipman2023flowmatching}. Exact integration from $\tau=1$
to $\tau=0$, with $P$ fixed, therefore transports Gaussian
continuation noise to the law of $(B_V)_{0:m}$.
Taking the first $s\le m$ actions yields
\begin{equation}
\pi_{\mathrm{RFD}}^\star(\cdot\mid o^-,P)
=
\mathbb E_{O^+\mid o^-,P}
\left[\pi_{\mathrm{VLA}}(\cdot\mid O^+)\right],
\end{equation}
where both policies denote distributions over the executed
$s$-step segment. This proves the proposition. \hfill$\square$

\paragraph{Relationship to teacher-sample supervision.}
Under the same population sampling law and MSE normalization,
teacher-sample supervision uses the target
$G:=(\epsilon-B_V)_{0:m}$ instead of $T_\tau$.
Equation~\ref{eq:exact-teacher-field} implies
$\mathbb E[G-T_\tau\mid Z_\tau,\tau,O^+]=0$.
Because $w(X_\tau,\tau)$ is determined by $Z_\tau$ and the fixed
early context, expanding the squared error makes the cross
term vanish, giving
\begin{equation}
\mathcal L_{\mathrm{sample}}(w)
=
\mathcal L_{\mathrm{RFD}}(w)
+
\mathbb E\!\left[\operatorname{MSE}(G,T_\tau)\right].
\label{eq:sample-field-loss-relation}
\end{equation}
The final term is independent of the student, so the two
objectives have the same population minimizers. RFD replaces
a sample-derived velocity label with the teacher's conditional
mean velocity; it changes the supervision, not the ideal target.
These identities concern population sampling and an exact
teacher, rather than a guarantee for a fixed finite teacher
bank or finite-step implementation.

\clearpage
\section{Interpreting Gaussian Resolution}
\label{app:resolver-analysis}

\paragraph{Moment-matched Gaussian approximation.}
Fix a handoff observation $o^+$ and write
$p(U):=\pi_{\mathrm{VLA}}(U\mid o^+)$, with each continuation
represented by $n$ scalar coordinates. Assume finite means
$\mu_j=\mathbb E_p[U_j]$ and positive finite variances
$\sigma_j^2=\operatorname{Var}_p(U_j)$.
For a diagonal Gaussian
$q_{m,v}=\mathcal N(m,\operatorname{diag}v)$ with $v_j>0$,
its expected negative log density is
\begin{equation}
\mathbb E_p[-\log q_{m,v}(U)]
=
\frac12\sum_{j=1}^{n}
\left[
\log(2\pi v_j)
+
\frac{\sigma_j^2+(\mu_j-m_j)^2}{v_j}
\right].
\end{equation}
Minimizing coordinate-wise gives $m_j=\mu_j$ and
$v_j=\sigma_j^2$. Thus, whenever the forward KL divergence
is finite,
\begin{equation}
q^\star
=
\arg\min_{q\in\mathcal G_{\mathrm{diag}}}
D_{\mathrm{KL}}(p\Vert q)
=
\mathcal N(\mu,\operatorname{diag}\sigma^2).
\label{eq:gaussian-forward-kl}
\end{equation}
This is the standard moment-matching characterization of
Gaussian approximation~\citep{minka2001expectation}.
The empirical moments estimated from $M$ teacher samples
in Section~\ref{sec:resolution} approximate these population
parameters, and the learned predictor approximates their
dependence on $o^+$. This justifies the moment targets;
it does not assume that $p$ is Gaussian or that the learned
$\hat\pi_\phi$ equals $q^\star$ exactly.

\paragraph{What the Gaussian ranking measures.}
For a fixed handoff observation, the Gaussian normalizing
factor is identical across candidates. Hence,
\begin{equation}
\arg\max_k q^\star(U^k)
=
\arg\min_k
\sum_{j=1}^{n}
\frac{(U^k_j-\mu_j)^2}{\sigma_j^2}.
\label{eq:gaussian-distance-ranking}
\end{equation}
Deviations along coordinates where the teacher varies little
are penalized more strongly than deviations along coordinates
where it varies substantially. This score also has an
interpretation directly under the original teacher, without
assuming Gaussianity. For any fixed candidate $u$ and
an independent teacher continuation $A\sim p$,
\begin{equation}
\mathbb E_{A\sim p}\!\left[
\sum_{j=1}^{n}\frac{(u_j-A_j)^2}{\sigma_j^2}
\right]
=
\sum_{j=1}^{n}\frac{(u_j-\mu_j)^2}{\sigma_j^2}+n.
\label{eq:expected-standardized-discrepancy}
\end{equation}
To obtain this identity, expand
$u_j-A_j=(u_j-\mu_j)-(A_j-\mu_j)$: the cross term has
zero expectation, and the remaining variance term contributes
one per coordinate. Gaussian ranking therefore selects the
available candidate minimizing expected variance-normalized
squared discrepancy to a teacher draw. This interpretation
requires only the stated moment assumptions, not a unimodal
teacher distribution.

\paragraph{A multimodal example.}
Consider the two-dimensional teacher
\begin{equation}
p
=
\tfrac12\mathcal N((-1,1),\eta^2 I)
+
\tfrac12\mathcal N((1,1),\eta^2 I),
\qquad 0<\eta\ll1.
\label{eq:gaussian-resolution-example}
\end{equation}
Its mean is $(0,1)$ and its coordinate variances are
$(1+\eta^2,\eta^2)$. Suppose the candidate bank contains
$(-1,1)$, $(1,1)$, $(-1,-1)$, and $(1,-1)$.
The first coordinate separates the teacher's two modes,
whereas the second is consistently near $1$.
Equation~\ref{eq:gaussian-distance-ranking} assigns the latter
two candidates an additional penalty of $4/\eta^2$,
strongly preferring either of the first two.
The Gaussian need not represent the two modes separately
to reject candidates that violate their shared second-coordinate
behavior. Moreover, the resolver returns an existing
candidate; it does not synthesize or execute the mean $(0,1)$.

\paragraph{Limitations and interpretation.}
The same example shows where this reasoning can fail.
If $(0,1)$ is added to the candidate bank, the Gaussian ranks
it above both component centers, although its teacher density
is very small for small $\eta$. Selection avoids creating
an averaged trajectory, but does not prevent selecting an
undesirable intermediate candidate already present in the bank.
Diagonal covariance also discards correlations across action
coordinates and timesteps, and distributions with identical
coordinate-wise moments receive identical Gaussian summaries.
Consequently, Gaussian ranking need not agree with the ideal
VLA-likelihood ranking in Eq.~\ref{eq:resolve-ideal}, reconstruct
the teacher distribution, or preserve its mode probabilities.
Finite-sample moment estimates and imperfect parameter
prediction introduce further approximation.

The resulting division of labor is therefore deliberate:
the proposer supplies structured alternative continuations,
while the resolver uses the handoff observation to compare
them through a cheap moment-based score. The score need not
serve as a faithful generative policy to be useful for this
selection problem; whether its ranking is sufficient is
evaluated through the closed-loop experiments.

\section{Additional Distribution Analysis}
\label{sec:appendix}
\label{sec:fielddistill-insights}
\label{app:coverage}

\paragraph{Support metric.}
For each handoff state, the VLA provides 256 action samples, split into 128 support samples and 128 independent query samples; each asynchronous method provides 128 proposals. Following \citet{kynkaanniemi2019precision}, we estimate the support of a sample set as the union of balls centered at its samples, with each radius given by the distance to that sample's fifth nearest neighbor within the set, excluding itself. Distances are computed over the policy-normalized executed action segment as
\[
d(A,B)=\frac{\lVert\operatorname{vec}(A-B)\rVert_2}{\sqrt{sD_a}},
\]
using only physical action coordinates. Precision is the fraction of asynchronous proposals contained in the estimated VLA support, while recall is the fraction of independent VLA queries contained in the estimated asynchronous support. Figures~\ref{fig:handoff-options} and~\ref{fig:fidelity-histograms} report these raw quantities without normalization. VLA self-coverage is obtained by testing the VLA query bank against the VLA support bank. The supplementary plots in Figures~\ref{fig:coverage-delay} and~\ref{fig:teacher-samples} report relative precision and recall, dividing both raw metrics independently at each handoff state by that state's VLA self-coverage; these normalized values may exceed one.

\paragraph{VLA self-coverage reference.}
At $d=10$, using the same states and equal task weighting as Figure~\ref{fig:fidelity-histograms}, mean VLA self-coverage is $0.8900$ on demonstration states and $0.8836$ on rollout states for RoboMimic, and $0.8578$ and $0.8592$, respectively, for LIBERO. These averages first average self-coverage within each task, then equally weight the three RoboMimic tasks or 40 LIBERO tasks. For the ToolHang histogram population, mean self-coverage is $0.8954$ ($0.8906$ for RTC's smaller state subset).

\subsection{Generalization Across Delay}
\label{app:coverage-delay}
\label{app:delay-fidelity}
\label{app:coverage-generality}

\begin{figure}[!htbp]
  \centering
  \includegraphics[width=\linewidth]{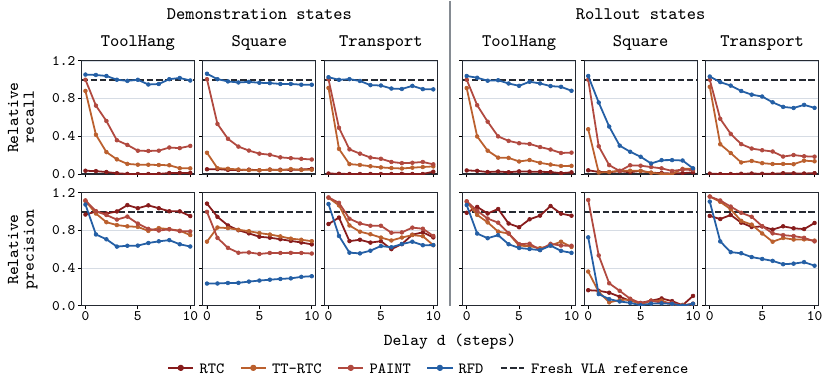}
  \setlength{\abovecaptionskip}{4pt}
  \caption{\textbf{Distribution alignment across delays.}
  The support comparison from Figure~\ref{fig:fidelity-histograms}, extended across delays $d=0,\ldots,10$ on ToolHang, Square, and Transport, showing how relative recall and precision evolve on demonstration and rollout states.}
  \label{fig:coverage-delay}
  \label{fig:fidelity-comparison}
\end{figure}

Figure~\ref{fig:coverage-delay} extends the distribution analysis across inference delays. On demonstration states, RFD maintains high recall across delays, indicating that the learned objective remains robust as the committed prefix grows. We additionally evaluate states encountered during VLA rollouts, where RFD must generalize to observations and prefixes not seen during training. This generalization remains strong on ToolHang and degrades only moderately on Transport, but drops substantially with increasing delay on Square. Thus, the main limitation in proposal coverage appears when predicting the handoff-time VLA behavior from out-of-distribution rollout states.

\FloatBarrier

\subsection{Teacher-Sample Supervision}
\label{app:teacher-samples}
\label{app:teacher-sample-distribution}

\begin{figure}[!t]
  \centering
  \includegraphics{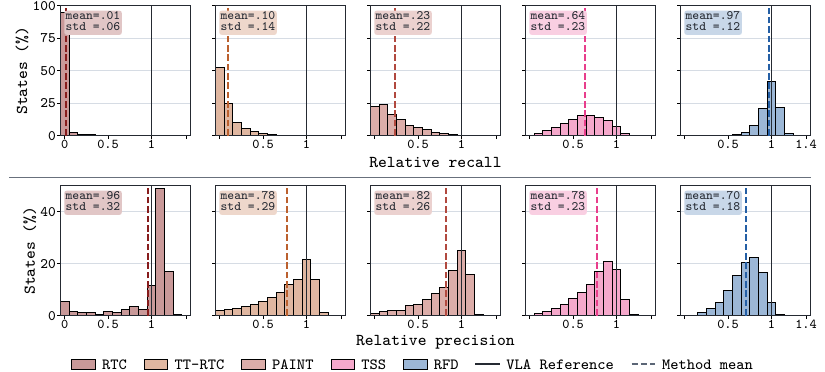}
  \setlength{\abovecaptionskip}{4pt}
  \caption{\textbf{Action support with teacher-sample supervision.}
  The ToolHang comparison from Figure~\ref{fig:fidelity-histograms}, now including teacher-sample supervision (TSS) and reporting precision and recall relative to each state's VLA self-coverage.}
  \label{fig:teacher-samples}
  \label{fig:teacher-sample-support}
\end{figure}

Figure~\ref{fig:teacher-samples} compares RFD with the teacher-sample supervision introduced in Section~\ref{sec:flow-field-distillation}, which replaces demonstration continuations with samples from the VLA at $o^+$ but trains on the sampled endpoints rather than the teacher flow field. Teacher-sample supervision improves substantially over the original continuation objectives, but does not recover the VLA support as completely as RFD. We tested supervision with $K\in\{1,8,16\}$ teacher samples per condition and observed the same qualitative limitation, whereas RFD achieved near-complete recall directly. Although both objectives target the same teacher distribution in the population limit, transferring the teacher flow field provides a clear empirical advantage for learning that distribution in practice. We therefore use RFD as the proposer throughout the remaining experiments.

\FloatBarrier

\section{Implementation Details}
\label{app:implementation-details}

\subsection{Setup}
\label{app:setup}

For RoboMimic, we initialize from \texttt{lerobot/smolvla\_base} and fine-tune one SmolVLA policy per task (Square, Transport, ToolHang), training the action expert and projection layers while keeping the vision-language backbone frozen. We use the selected 60k-update checkpoints for all three tasks. For LIBERO, we use the released \texttt{HuggingFaceVLA/smolvla\_libero} checkpoint directly and train only our additional asynchronous components. RTC, TT-RTC, and PAINT are implementations of the published algorithms adapted to the corresponding backbone and execution protocol.

\noindent\textbf{Evaluation protocol.} All policies use action horizon $H=50$, with ten Euler steps per flow integration. We execute $s=10$ actions per chunk on RoboMimic and $s=25$ on LIBERO. On RoboMimic, we evaluate 100 episodes each on Square and Transport and 200 on ToolHang; the reported average weights the three task success rates equally. On LIBERO, we evaluate all 40 tasks with 50 episodes per task, for 2,000 episodes per condition. Initial states are shared across methods, delays, and the synchronous VLA reference. We evaluate $d\in\{5,\ldots,10\}$ on RoboMimic and $d\in\{5,10,15,20\}$ on LIBERO.

\paragraph{Training configuration.}
Tables~\ref{tab:optimization-settings} and~\ref{tab:training-configurations} report the optimization settings and training seeds for the checkpoints used in the closed-loop experiments. Each component trained in this work uses a single training seed.

\begin{center}
\begin{minipage}{\linewidth}
  \centering
  \small
  \captionof{table}{\textbf{Optimization settings.} Shared settings for base-VLA fine-tuning, TT-RTC, RFD, and resolver training.}
  \label{tab:optimization-settings}
  \begin{tabular}{@{}lcc@{}}
    \toprule
    Hyperparameter & Base VLA / TT-RTC / RFD & Resolver \\
    \midrule
    Optimizer & AdamW & AdamW \\
    Adam $(\beta_1,\beta_2)$ & $(0.9,0.95)$ & $(0.9,0.95)$ \\
    Weight decay & $10^{-10}$ & $10^{-4}$ \\
    Gradient clipping norm & 10 & 5 \\
    Learning-rate schedule & Cosine with warm-up & Cosine with warm-up \\
    Warm-up steps & 1,000 & 500 \\
    Terminal learning rate & $2.5\times10^{-6}$ & 5\% of peak \\
    \bottomrule
  \end{tabular}
\end{minipage}
\end{center}

\begin{center}
\begin{minipage}{\linewidth}
  \centering
  \small
  \setlength{\tabcolsep}{4pt}
  \captionof{table}{\textbf{Training configurations.} Policy update counts extend through the evaluated checkpoint; resolver counts give the full training budget. Learning rates are configured peaks. Arrows and sums denote successive stages.}
  \label{tab:training-configurations}
  \begin{tabular}{@{}llccrr@{}}
    \toprule
    Component & Task & Peak LR & Batch & Updates & Seed \\
    \midrule
    Base VLA & Square & $10^{-4}$ & 16 & 60,000 & 1000 \\
    Base VLA & Transport & $10^{-4}$ & 8 & 60,000 & 1000 \\
    Base VLA & ToolHang & $10^{-4}$ & 16 & 60,000 & 1000 \\
    Base VLA & LIBERO & --- & --- & Released checkpoint & --- \\
    \midrule
    TT-RTC & Square & $10^{-4}$ & 16 & 100,000 & 1000 \\
    TT-RTC & Transport & $2.5\times10^{-5}$ & 16 & 25,000 & 1000 \\
    TT-RTC & ToolHang & $10^{-4}\to2.5\times10^{-5}$ & $16\to64$ & $10{,}000+25{,}000$ & 1000 \\
    TT-RTC & LIBERO & $2.5\times10^{-5}$ & 16 & 25,000 & 20260921 \\
    \midrule
    RFD & Square & $10^{-4}$ & 16 & 100,000 & 1000 \\
    RFD & Transport & $10^{-4}\to2.5\times10^{-5}$ & $8\to16$ & $50{,}000+16{,}000$ & 1000 \\
    RFD & ToolHang & $10^{-4}\to2.5\times10^{-5}$ & $16\to64$ & $50{,}000+25{,}000$ & 1000 \\
    RFD & LIBERO & $2.5\times10^{-5}$ & 16 & 25,000 & 20260921 \\
    \midrule
    Resolver & Square & $3\times10^{-4}$ & 16 & 240,000 & 20260825 \\
    Resolver & Transport & $3\times10^{-4}$ & 16 & 240,000 & 20260825 \\
    Resolver & ToolHang & $3\times10^{-4}$ & 16 & 240,000 & 20260903 \\
    Resolver & LIBERO & $10^{-4}$ & 256 & 10,000 & 20260921 \\
    \bottomrule
  \end{tabular}
\end{minipage}
\end{center}

\paragraph{Training stages.}
TT-RTC and RFD update counts exclude preceding base-VLA training. ToolHang TT-RTC and RFD resume the 50k base checkpoint with optimizer state before their final fresh-optimizer stage; Transport RFD likewise uses a fresh optimizer for its final stage. The ToolHang base VLA and initial continuation stages retain the original 100k-step schedule; Square RFD reaches its terminal learning rate at 50k updates and retains it thereafter. TT-RTC uses the final checkpoint on all tasks; RFD uses the final checkpoint except on Square, where the 100k checkpoint is selected from an offline precision/recall sweep extending to 200k updates.

\paragraph{Baseline inference.}
RTC uses ten Euler steps, the EXP prefix-weight schedule, a maximum guidance weight of 10, and guidance through the full predicted-endpoint Jacobian. Its guidance horizon is $H-s$: 40 actions on RoboMimic and 25 on LIBERO. PAINT uses one repainting cycle: a ten-step generation pass, a ten-step inversion of the generated chunk with its first $d$ actions replaced by the committed prefix, and a ten-step regeneration pass. The regeneration noise combines the inverted prefix noise with the original suffix noise. Thus, PAINT uses 30 flow evaluations per proposal at positive delay. All three passes use the observation available when asynchronous inference begins.

\subsection{Recursive Flow-Field Distillation}
\label{app:rfd-training}

For each logged observation, we precompute $8$ continuations from the frozen VLA. During RFD training, one teacher continuation is sampled, combined with fresh Gaussian noise at a sampled flow time, and used to query the frozen VLA's velocity field at the reached observation $o^+$. The student receives the earlier observation $o^-$ together with the committed prefix $P$, and is trained with the delay-aligned velocity-matching objective from Section~\ref{sec:flow-field-distillation}.

\subsection{Propose--Resolve Implementation}
\label{app:resolver-implementation}

The resolver is trained from frozen-VLA action samples and frozen action-expert features. For each training observation $o^+$, we draw $M=32$ VLA continuations and compute their empirical per-coordinate mean and standard deviation over the resolved action segment. A lightweight MLP predicts the corresponding mean and log-standard-deviation from frozen VLA features, using robust Smooth-L1 regression. At inference, features are computed once from the newly observed $o^+$, and the resulting diagonal Gaussian scores all $K$ proposed continuations by their log density. PR executes the highest-scoring candidate. The VLA remains frozen throughout resolver training, and the same resolver can be applied to proposals from RFD or any of the asynchronous baselines.

\paragraph{Architecture and checkpoint selection.}
The resolver is an MLP with hidden widths 754, 512, and 256. Each hidden linear layer is followed by LayerNorm and GELU; dropout is zero. Separate linear heads predict the mean and log-standard-deviation over the executed action segment. We select resolver checkpoints by minimum validation negative log-likelihood per action coordinate, using episode-balanced averaging on ToolHang. Validation is performed every 5,000 updates on RoboMimic and every 500 updates on LIBERO. The selected checkpoints are at 225,000, 140,000, and 70,000 updates for Square, Transport, and ToolHang, respectively, and 4,500 updates for LIBERO.

\subsection{Timing Attribution}
\label{app:timing}

Propose--Resolve separates computation into two stages with different timing requirements. Proposal generation occurs while the committed prefix is being executed and therefore has a budget of $d$ control steps. Resolution begins only once the handoff observation $o^+$ becomes available and lies directly on the control-critical path.

\begin{table}[t]
  \centering
  \caption{\textbf{Compute latency of Propose--Resolve.}
  RFD proposals with energy-argmax resolution on an NVIDIA GeForce RTX~4090, using 50-action chunks and ten flow steps. All times are in milliseconds. Proposal generation overlaps with execution; resolution follows observation of $o^+$. At $K=1$, resolution is bypassed. Total means sum separately measured proposal and resolution means; they are not joint pipeline measurements.}
  \label{tab:timing-appendix}
  \begin{tabular}{lrrrr}
    \toprule
    Model & $K$ & Propose mean / p99 & Resolve mean / p99 & Total mean \\
    \midrule
    SmolVLA & 1 & 27.0 / 27.6 & 0 / 0 & \textbf{27.0} \\
    SmolVLA & 8 & 39.8 / 40.4 & 12.0 / 12.4 & \textbf{51.8} \\
    SmolVLA & 16 & 56.5 / 57.1 & 12.0 / 12.5 & \textbf{68.6} \\
    SmolVLA & 32 & 90.8 / 91.4 & 12.0 / 12.3 & \textbf{102.8} \\
    \bottomrule
  \end{tabular}
\end{table}

Table~\ref{tab:timing-appendix} reports the measurements across candidate counts. For SmolVLA, preparing $K=32$ alternatives takes $90.8$\,ms on average ($91.4$\,ms at p99), while resolution takes $12.0$\,ms on average ($12.3$\,ms at p99). Thus, resolution remains within a $50$\,ms control period at p99, while the more expensive proposal generation occurs before handoff within the asynchronous execution window.

Measurements use one NVIDIA GeForce RTX~4090, with RFD proposals, energy-argmax resolution, 50-action chunks, and ten flow steps. SmolVLA uses two $512\times512$ camera images with BF16 execution. Generation is measured at $d=5$ and resolution at $d=10$. Proposal and resolution are benchmarked separately and include preprocessing, CPU--GPU transfers, model computation, and CPU outputs; camera acquisition and robot I/O are excluded. Proposal generation shares the encoded observation context across candidates, while resolution computes the fresh-observation features once and scores the entire candidate bank jointly.

% Timing sources supplied with these measurements:
% /home/zoellner/src/propose-resolve/outputs/latency/20260923-smol-generation01/benchmark/README.md
% /home/zoellner/src/vla-rl-robosuite/outputs/benchmarks/rtc_inference/20260909_k_4090/analysis/README.md

\end{document}